%% file: main.tex
\documentclass{mlrpreprint}

\input{preamble}

\newcommand*\methodname{PointAction\xspace}

\crefname{section}{Sec.}{Secs.}
\Crefname{section}{Section}{Sections}
\crefname{subsection}{Sec.}{Secs.}
\Crefname{subsection}{Section}{Sections}
\crefname{figure}{Fig.}{Figs.}
\Crefname{figure}{Figure}{Figures}
\crefname{table}{Table}{Tables}
\Crefname{table}{Table}{Tables}
\crefname{equation}{Eq.}{Eqs.}
\Crefname{equation}{Equation}{Equations}
\crefname{algorithm}{Alg.}{Algs.}
\Crefname{algorithm}{Algorithm}{Algorithms}
\crefname{appendix}{Appendix}{Appendices}
\Crefname{appendix}{Appendix}{Appendices}

\title{PointAction: 3D Points as Universal Action Representations for Robot Control}

\author[\dagger]{Mutian Tong}
\author[\dagger*]{Han Jiang}
\author{Qiao Feng}
\author{Lingjie Liu}
\author{Jiatao Gu}

\affiliation{University of Pennsylvania}

\contribution[\dagger]{Equal contribution}

\abstract{
Video-Action Models (VAMs) leverage the broad visual dynamics captured by pre-trained video diffusion models, offering a promising path toward generalizable robot manipulation. However, RGB-only video rollouts are not directly actionable: they leave metric 3D motion, contact geometry, and fine-grained spatial constraints under-specified, making action grounding ambiguous. Meanwhile, scaling action supervision across diverse tasks and embodiments remains costly. We present \textbf{\methodname{}}, a framework that bridges video predictions to robot actions through explicit point-based 4D modeling. \methodname{} fine-tunes a foundation video generation model to jointly predict future RGB frames and dynamic 3D pointmaps, producing temporally consistent 3D motion of task-relevant scene geometry. These point dynamics serve as a structured, embodiment-agnostic action interface, which a diffusion-based action decoder maps to executable robot actions. By using metric 3D point dynamics as the interface between video prediction and control, \methodname{} reduces the ambiguity of RGB-only action grounding and supports transfer across tasks and embodiments with limited action supervision. Experiments show that \methodname{} achieves state-of-the-art 4D generation quality on robot scenes, outperforms existing baselines in simulation, and generalizes to two real robot arms unseen during pretraining.
}

\correspondence{\email{jgu32@seas.upenn.edu}}
\newcommand\projectpage[1]{\metadata[\bsicon{\faGlobe}\,\textbf{Project page:}]{#1}}
\projectpage{\url{https://oriontmt.github.io/pointaction/}}
\begin{document}
\maketitle
\blfootnote{Work done during internship at University of Pennsylvania.}

%======================================================================
% Teaser figure: rendered immediately under the title/abstract block.
%======================================================================
\begin{figure}[!h]
  \centering
  \includegraphics[width=1.0\textwidth]{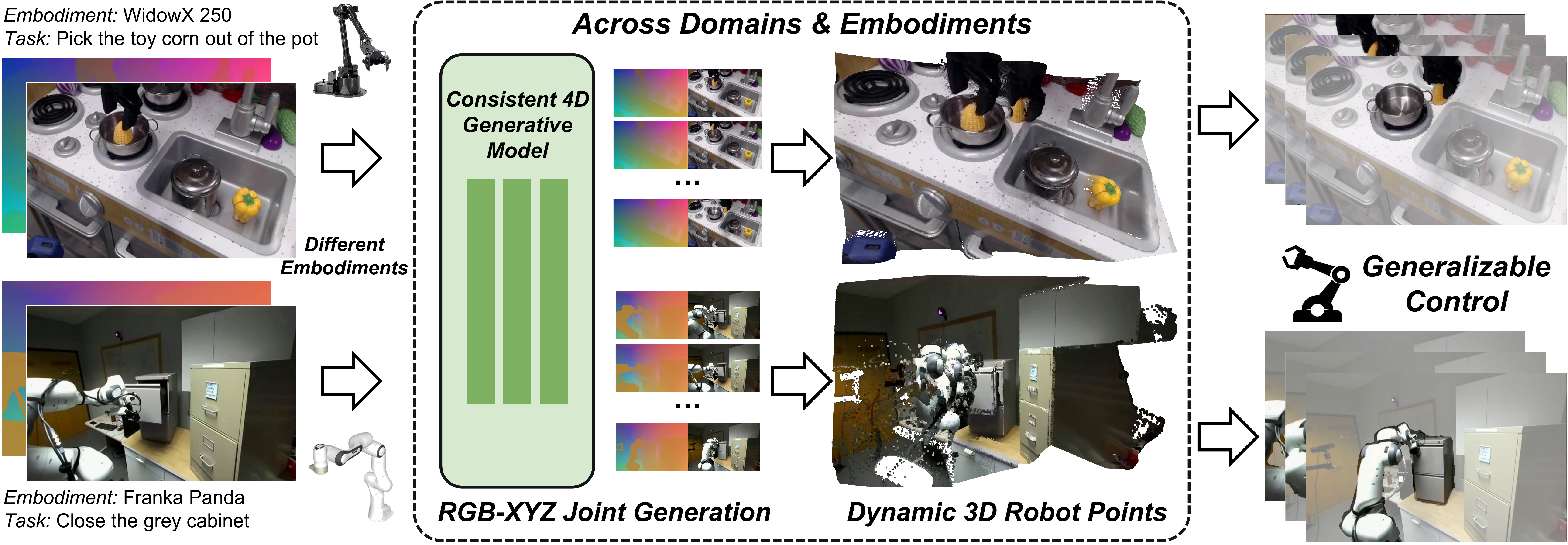}
  \caption{\small\textbf{\methodname enables generalizable robot control via explicit 4D point modeling.} From an input image and instruction, our framework jointly synthesizes consistent RGB frames and dynamic, pixel-aligned 3D pointmaps (XYZ). These dynamic 3D robot points serve as a generalizable and scalable proxy for action decoding, seamlessly bridging 2D video priors to robust visuomotor control.}
  \label{fig:overview}
\end{figure}

\input{sections/02_intro}
\input{sections/03_relatedworks2}
\input{sections/04_methods}

\input{sections/05_experiments}
\input{sections/06_disucssion}

%======================================================================
\bibliographystyle{plainnat}
\bibliography{main}

\clearpage
\appendix
\input{sections/07_appendix}

\end{document}

%% file: preamble.tex
\usepackage{amsmath}
\usepackage{enumerate}
\usepackage{algorithm}
\usepackage{algpseudocode}
\usepackage{amsfonts}
\usepackage{amsthm}
\usepackage{cleveref}
\usepackage{diagbox}
\usepackage{colortbl}
\usepackage{amssymb}
\usepackage{xspace}
\usepackage{wrapfig}
\usepackage{adjustbox}
\usepackage{tabularx}
\usepackage{booktabs}
\usepackage{mathtools}
\usepackage{tikz}
\usepackage{enumitem}
\usepackage{silence}
\usepackage{dsfont}
\usepackage{multirow}
\usepackage{makecell}
\usepackage{xfakebold}
\usepackage{listings}
\usepackage{nicefrac}
\usepackage{url}
\usepackage{fontawesome5}
\usepackage[cal=boondoxo]{mathalfa}
\input{math_commands}

\definecolor{textgray}{HTML}{6E6E73}
\usetikzlibrary{positioning, calc}
\usetikzlibrary{decorations.pathmorphing}

\makeatletter
\patchcmd{\wrong@fontshape}{\@gobbletwo}{}{}{}
\makeatother
\numberwithin{equation}{section}
\makeatletter
\AtBeginDocument{
  \urlstyle{sf}
  
}
\makeatother

\definecolor{tabfirst}{rgb}{1, 0.7, 0.7}
\definecolor{tabsecond}{rgb}{1, 0.85, 0.7}
\definecolor{tabthird}{rgb}{1, 1, 0.7}
\definecolor{tabgray}{rgb}{0.9, 0.9, 0.9}
\definecolor{turquoise}{cmyk}{0.65,0,0.1,0.3}
\definecolor{darkgreen}{rgb}{0, 0.5, 0}
\definecolor{darkred}{rgb}{0.6, 0.1, 0.05}
\definecolor{blueish}{rgb}{0.0, 0.3, .6}
\definecolor{light_gray}{rgb}{0.7, 0.7, .7}
\definecolor{greyblue}{rgb}{0.25, 0.25, 1}

\makeatletter
\def\onedot{\futurelet\@let@token\@onedot}
\def\@onedot{\ifx\@let@token.\else.\null\fi\xspace}
\makeatother

\newcommand\rgt{\aftergroup\mathclose\aftergroup{\aftergroup}\right}

\makeatletter
\newcommand\blfootnote[1]{%
  \begingroup
    \def\@thefnmark{*}%
    \begin{NoHyper}\@footnotetext{#1}\end{NoHyper}%
  \endgroup
}
\makeatother

%% file: math_commands.tex
\usepackage{amsmath,amsfonts,bm}

\def\eqref#1{equation~\ref{#1}}
\def\1{\bm{1}}

\DeclareMathAlphabet{\mathsfit}{\encodingdefault}{\sfdefault}{m}{sl}
\SetMathAlphabet{\mathsfit}{bold}{\encodingdefault}{\sfdefault}{bx}{n}

%% file: sections/02_intro.tex
\section{Introduction}
Building generalist robot policies capable of solving diverse manipulation tasks across environments and embodiments remains a central goal in robot learning.
Recent Vision-Language-Action (VLA) models~\cite{vla_survey_1,vla_survey_2,rt2,openvla,pi0,pi05,gr00t} have made remarkable progress by combining the semantic priors of large pre-trained Vision-Language Models (VLMs)~\cite{team2024gemma,bai2025qwen3} with large-scale robot datasets~\cite{oxe,droid,bridgev2}.
These models can follow language instructions and recognize task-relevant objects with increasing robustness.
However, semantic visual understanding alone does not provide an explicit model of how scenes evolve through contact, motion, and long-horizon interaction.
As a result, policies trained primarily from paired observation--action data can struggle when manipulation requires reasoning about future physical changes beyond the training distribution.

To incorporate such dynamics, recent work has explored Video-Action Models (VAMs)~\cite{unipi,dreamgen,susie,mimic-video,linbot-va,cosmos-policy}, which use pretrained video diffusion backbones~\cite{cosmos-predict2,wan,cogvideox,wiedemer2025video} to predict future scene rollouts as an explicit reasoning trace for action generation.
Existing methods either jointly model future observations and actions in an end-to-end process~\cite{linbot-va,dreamzero}, or first predict future observations and then decode actions through an inverse-dynamics module~\cite{unipi,dreamgen,susie,mimic-video}.
Despite their promise, current VAMs still face two tightly coupled obstacles.
First, even when actions are predicted jointly or decoded from rollouts, the intermediate representation is typically RGB-dominant: 3D motion, contact-relevant geometry, and fine-grained spatial constraints remain implicit, forcing the action module to learn a difficult mapping from appearance changes to controls.
Second, learning this implicit grounding requires substantial paired observation--action supervision~\cite{gofe}; however, such data is costly to collect, tied to specific robot embodiments, and difficult to scale across diverse tasks and environments.

We argue that dynamic 3D pointmaps provide a natural way to break this representation--supervision bottleneck. As an intermediate representation, they make predicted rollouts more actionable by exposing how task-relevant 3D geometry moves over time, including metric motion and contact-relevant spatial constraints that RGB frames leave implicit. As a training signal, point-level supervision can be obtained at scale from videos through multi-view reconstruction, monocular depth estimation, and motion cues, allowing the rollout model to learn broad scene dynamics beyond robot-specific action labels. Building on this insight, we propose \methodname, a framework that connects video prediction to robot control through explicit point-based 4D modeling.

Concretely, \methodname factorizes video-to-action learning into two components: (1) a universal video-to-point model that predicts RGB rollouts together with dynamic 3D pointmaps, and can be pretrained on large-scale video data across tasks and embodiments; and (2) an embodiment-specific point-to-action decoder that maps predicted point dynamics to executable controls and can be adapted with limited robot data. We implement the universal component by fine-tuning a foundation robot video model~\cite{lvp} to jointly generate RGB frames and spatially aligned XYZ pointmaps, and train a diffusion-based action decoder for action translation. Comprehensive evaluations show that \methodname achieves state-of-the-art 4D generation quality on robot scenes, outperforms existing VLA and VAM baselines in simulation, and transfers to real robot embodiments unseen during pretraining. Our main contributions are threefold:

\begin{itemize}[leftmargin=*]
    \item \textbf{Point-based interface for video-to-action learning.}
    We introduce \methodname, which uses dynamic 3D pointmaps as an explicit interface between video prediction and robot control, reducing RGB-only grounding ambiguity.

    \item \textbf{Scalable pretraining with embodiment-specific decoding.}
    \methodname separates a universal video-to-point model, pretrainable from large-scale video data, from a lightweight point-to-action decoder that adapts to each robot with limited action supervision.

    \item \textbf{Strong 4D generation and manipulation results.}
    Experiments show state-of-the-art 4D generation quality on robot scenes, improved simulation performance over VLA and VAM baselines, and transfer to real robot embodiments unseen during pretraining.
\end{itemize}

%% file: sections/03_relatedworks2.tex
\section{Related Works}

\noindent\textbf{Vision-Language-Action Models.}
Vision-Language-Action (VLA) models~\cite{vla_survey_1, vla_survey_2} achieve strong manipulation performance by leveraging pretrained vision-language backbones and fine-tuning on large-scale robot datasets~\cite{oxe, droid, bridgev2}.
RT-2~\cite{rt2} co-trained VLMs on robotic trajectories and web-scale vision-language data, and subsequent work~\cite{openvla, octo, pi0, pi05, gr00t} improves performance via larger datasets and better architectures.
A key gap is the insufficient modeling of world dynamics.
Recent VLAs incorporate visual-guided action prediction via intermediate future-image rollouts:
CoT-VLA~\cite{cot-vla} autoregressively predicts subgoal images as explicit visual CoT; UniVLA~\cite{univla} and GR-series~\cite{gr-1, gr-2} pretrain on future images; WorldVLA~\cite{worldvla} and RynnVLA-002~\cite{rynnvla-002} jointly optimize future image generation and action prediction; DreamVLA and 3DVLA~\cite{3dvla} further incorporate additional modalities such as depth.
In contrast to these VLM-centric approaches, we build on a pretrained video diffusion model~\cite{lvp} to better capture temporal and fine-grained dynamics.

\noindent\textbf{Video-Action Models.}
Video-Action Models (VAMs) leverage pretrained video generators for robot control and fall into two paradigms. \emph{Decoupled VAMs}~\cite{unipi,dreamgen,susie,vpp,videopolicy,mimic-video,linbot-va,cosmos-predict2,wan} fine-tune a video model on robotic data and decode actions via an inverse-dynamics module, with VPP~\cite{vpp} and VideoPolicy~\cite{videopolicy} extracting actions from the video latent for robustness. \emph{End-to-end VAMs}~\cite{pad,uva,UWM,videovla,cogvideox,cosmos-policy} jointly generate future frames and actions in a single process, e.g., Cosmos-Policy~\cite{cosmos-policy} encodes robot state, action, and value as pseudo video frames.

\noindent\textbf{4D Modeling for Robot Control.}
Early 4D generation methods relied on optimization with priors from video generators~\cite{poole2022dreamfusion, Wu_2024_CVPR, pumarola2020d, efficient4d, free4d, dreamgs4d, 4dfy}.
More recent feed-forward approaches~\cite{4dnex, wvd, jiang2025geo4d, bai2025geovideointroducinggeometricregularization, fang2025worldreel4dvideogeneration} use video diffusion models to synthesize 4D content, but primarily target static 3D or natural scenes and often miss fine-grained manipulation dynamics.
Concurrent work 4DGen~\cite{4dgen} adapts video diffusion to manipulation by jointly generating RGB and pointmaps, followed by FoundationPose~\cite{Wen2023FoundationPoseU6} for 6-DoF tracking; however, it requires a predefined gripper CAD model and assumes continuous end-effector visibility.
TesserAct~\cite{tesseract} predicts 2.5D outputs (RGB, depth, normals) and reconstructs 4D scenes via an additional normal-integration step, decoupling generation from visuomotor control.
By directly predicting spatially aligned 3D pointmaps (XYZ) in a single pass, our method provides a reconstruction-free geometric foundation that more seamlessly grounds 4D predictions into low-level robot actions.

%% file: sections/04_methods.tex
\section{\methodname}
\subsection{Preliminaries}
We study language-conditioned manipulation: at time $t$, the agent observes an RGB image $o_t \in\mathbb{R}^{H\times W\times 3}$ and a proprioceptive state $s_t$ under an instruction $l$, and must produce a $\Delta$-step action chunk $\tilde{a}\triangleq a_{t:t+\Delta-1}\in\mathbb{R}^{\Delta\times D}$. Video-action models (VAMs)~\cite{dreamzero,linbot-va,unipi} use a predicted future observation chunk $\tilde{o}\triangleq o_{t+1:t+\Delta}\in\mathbb{R}^{\Delta\times H\times W\times 3}$ as an explicit \emph{reasoning trace} for action generation, either by jointly modeling future observations and actions,
\begin{equation}
(\tilde{o}, \tilde{a}) \sim \pi^{\text{VAM}}_\theta(\cdot \mid s_t, o_t, l),
\label{eq.vam}
\end{equation}
or by first generating $\tilde{o}$ and then decoding actions through an inverse-dynamics model. When $\tilde{o}$ is RGB only, however, metric 3D motion and contact-relevant geometry remain implicit, making fine-grained action grounding ambiguous. \methodname mitigates this ambiguity by lifting the rollout from RGB frames to joint RGB-XYZ pointmaps.
\subsection{Method Overview}
\label{sec:overview}
In this paper, we propose \textbf{\methodname}, a new paradigm that separates video-based prediction from robot-specific control through dynamic 3D pointmaps. The key idea is that many manipulation tasks share task-level geometric structure across embodiments: objects must be pushed, grasped, lifted, or moved to desired configurations, even when the underlying robot morphologies and control spaces differ. Such structure is naturally expressed in 3D, where physical interactions are determined by how the robot and objects move, make contact, and satisfy spatial constraints in a shared metric world. Moreover, supervision for 3D point trajectories can be obtained at scale from videos, e.g., through multi-view reconstruction, monocular depth cues, and point tracking, allowing the rollout model to learn beyond robot-specific action labels.

In this work, we instantiate this shared variable as a robot-centric, pixel-aligned pointmap for translating predicted 3D motion into robot actions.
We denote $u_t\in\mathbb{R}^{H\times W\times 4}$, where each pixel stores a 3D coordinate and a binary robot mask,
$u_t(p)=(x_p,y_p,z_p,\alpha_p)$, with $\alpha_p=1$ if pixel $p$ lies on the robot surface and $\alpha_p=0$ otherwise.
The two channel groups are obtained from different parts of the pipeline (\cref{sec:vam}): the geometric channels $(x_p,y_p,z_p)$ are predicted jointly with the RGB rollout by the 4D video model, while the mask $\alpha$ is supplied at inference time by an auxiliary off-the-shelf segmentation pass over the predicted RGB trajectory.
We use this notation throughout and extract the robot-centric stream as $u_{\mathrm{robo}} = u_{xyz} \odot \alpha$ when needed.
With this representation, we decompose the video-action model through an explicit point-based latent action $\tilde{u}$:
\begin{equation}
\label{eq:pointaction}
\pi(\tilde{o},\tilde{a}\mid s_t,o_t,l)
\approx
\int
\pi^{\text{4DVM}}_\theta(\tilde{o}, \tilde{u}\mid o_t,l)
\;\cdot\;
\pi^{\text{DEC}}_\psi(\tilde{a}\mid \tilde{u},s_t)
\,d\tilde{u}.
\end{equation}
In practice, we approximate this marginalization by first sampling an RGB-XYZ rollout $(\tilde{o},\tilde{u})$ from the 4D video model and then decoding $\tilde{a}$ from the predicted point dynamics.
$\pi^{\text{4DVM}}_\theta$ is \emph{embodiment-agnostic} because it does not condition on the arm-specific state $s_t$ and predicts a shared point-based rollout $\tilde{u}$. The same pretrained 4D backbone can therefore be reused across downstream robots, while embodiment-specific adaptation is confined to the lightweight decoder rather than requiring a new video-model pretraining run. In principle, $\pi^{\text{4DVM}}_\theta$ can be trained on any video source with extractable RGB-XYZ supervision, such as multi-view reconstruction or monocular depth estimation. In this work, we train it on robot manipulation videos spanning two robot arms (\cref{sec:setup}) and show transfer to two additional arms unseen during pretraining (\cref{sec:realworld}). In contrast, $\pi^{\text{DEC}}_\psi$ takes the predicted point dynamics $\tilde{u}$ together with the arm-specific state $s_t$ and maps them to executable controls using a small amount of paired robot data. We describe these two components next.

\subsection{Universal Video-to-Action Pretraining}
\label{sec:vam}

\begin{figure}[t]
  \centering
  \includegraphics[width=1.0\textwidth]{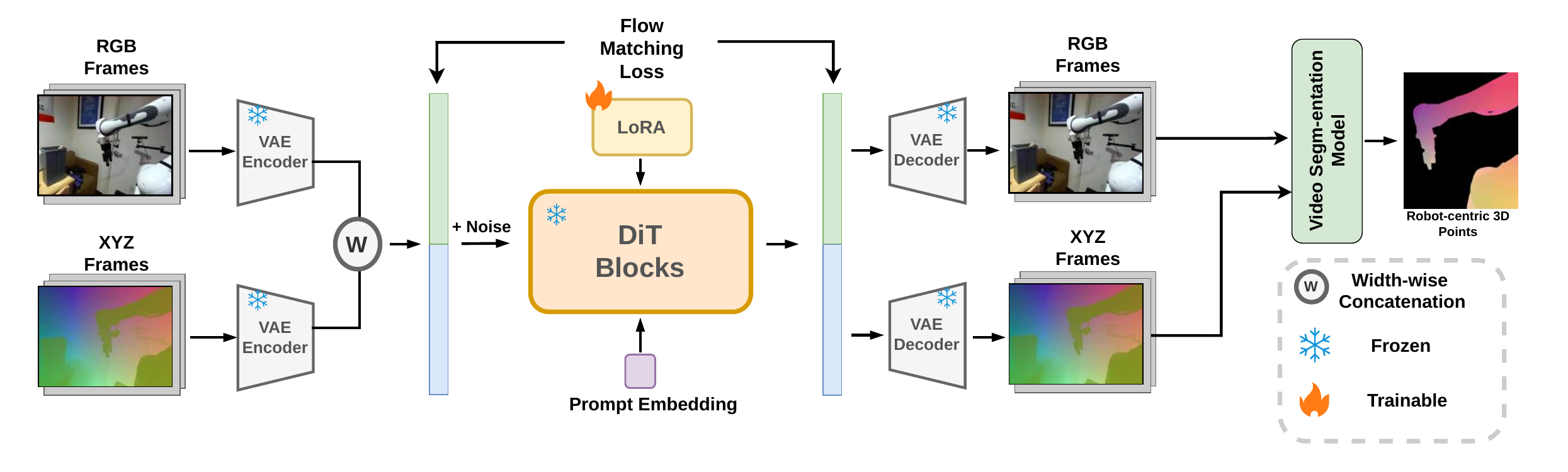}
  \caption{\small \textbf{Overview of the universal video-to-point model.}
We fine-tune a pretrained video model to jointly predict RGB rollouts and dynamic 3D pointmaps using the Diffusion Forcing objective.}
  \label{fig:stage1}
\end{figure}

To turn a video generation backbone into an action-grounded rollout model, we lift a 2D foundation video diffusion model to jointly predict RGB frames and dynamic 3D pointmaps. The goal is to preserve the broad visual dynamics learned from video pretraining while exposing metric 3D motion as an explicit interface for downstream control.

\noindent\textbf{Spatially Aligned Modality Fusion.}
A straightforward approach to joint RGB-XYZ generation is to treat the 3D pointmap $u$ as additional input channels, e.g., by using a 6-channel representation~\cite{wvd,jiang2025geo4d}. However, standard video DiTs pretrained on massive RGB data may struggle to align newly introduced geometric channels with their pretrained texture representations. Inspired by 4DNeX~\cite{4dnex}, we adopt a spatially aligned modality fusion strategy. Let $\mathcal{E}$ denote the frozen VAE encoder. We independently encode the RGB frames and XYZ pointmaps into the same latent space as $z^o = \mathcal{E}(o)$ and $z^u = \mathcal{E}(u)$, where $z \in \mathbb{R}^{C \times h \times w}$. We then concatenate the two latents along the spatial width dimension:
\begin{equation}
\tilde{z}^{\mathrm{joint}} = \mathrm{WidthConcat}(z^o, z^u) \in \mathbb{R}^{C \times h \times 2w}.
\end{equation}
This design preserves the pretrained channel structure of the RGB backbone while placing each visual patch close to its spatially corresponding geometric patch, allowing DiT self-attention to model local RGB-XYZ interactions without introducing new input channels.

\noindent\textbf{Training on Joint Latents.}
We initialize from LVP~\cite{lvp}, a foundation robot video model, and fine-tune the backbone on joint RGB-XYZ latent sequences. Following Diffusion Forcing~\cite{diffusionforcing,causvid}, each temporal sequence is randomly split into a history context $\hat z$ and a future trajectory $\tilde z$. The history context is assigned an independent noise level $\tau'$ and is set to be clean with $50\%$ probability, while the model predicts the continuous flow field of the future trajectory conditioned on the task instruction $l$. We train this joint rollout model with a Flow Matching objective~\cite{flow_matching}. Fine-tuning is performed parameter-efficiently with LoRA~\cite{lora}, which empirically preserves the backbone's RGB generation quality while adapting it to the joint RGB-XYZ target. The full training objective $\mathcal{L}_{\mathrm{flow}}$ and noise schedule are provided in \cref{sec:appendix_impl}.

\noindent\textbf{Extracting Robot-Centric Point Trajectories.}
As defined in \cref{sec:overview}, the action decoder operates on robot-centric point trajectories rather than the full scene-level pointmap.
Since the universal rollout model predicts dense XYZ pointmaps without embodiment-mask supervision, we obtain the mask $\tilde{\alpha}$ at inference time by applying an off-the-shelf video segmentation model, SAM~3~\cite{sam3}, to the predicted RGB trajectory $\tilde{o}$ with an open-vocabulary ``robot'' prompt.
The decoder input is then given by $\tilde{u}_{\mathrm{robo}}=\tilde{u}_{xyz}\odot\tilde{\alpha}$, where the mask is broadcast over the three geometric channels.

\begin{figure}[t]
  \centering
  \includegraphics[width=0.8\textwidth]{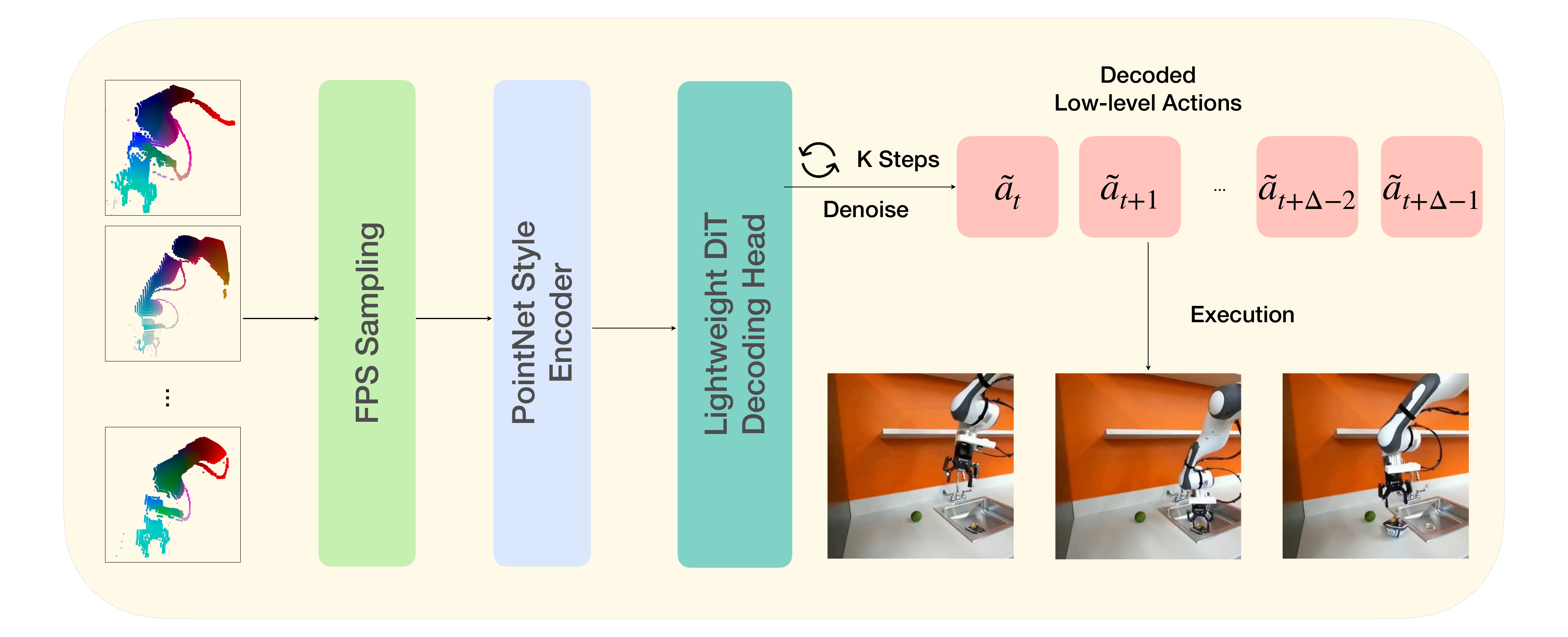}
  \caption{\small \textbf{Overview of the point-to-action decoder.} Given the predicted robot-centric pointmaps, we downsample each frame via Farthest Point Sampling and encode it with a lightweight PointNet-style MLP. A thin conditional DiT decoder then denoises the full action sequence in parallel, taking the per-frame point features as token-aligned conditioning and injecting the diffusion timestep and the initial robot state via AdaLN.}
  \label{fig:stage2}
\end{figure}

\subsection{Embodiment-Specific Point-to-Action Decoding}
The universal video-to-point model predicts how task-relevant 3D geometry should evolve, but these point dynamics must still be translated into the low-level action space of a particular robot. We therefore use an embodiment-specific point-to-action decoder that maps predicted robot-centric point trajectories to action chunks. This design keeps the expensive rollout model shared across tasks and embodiments, while requiring paired supervision only for the smaller action decoder.

Concretely, given the robot-centric point trajectories $\tilde{u}_{\mathrm{robo}}$ extracted above, we train a lightweight conditional diffusion model inspired by 3D Diffusion Policy~\cite{Ze2024DP3}. For each predicted frame, we downsample the point set to $N{=}512$ points using Farthest Point Sampling (FPS)~\cite{Qi2016PointNetDL} and encode the sampled points with a PointNet-style MLP $\Phi$. A lightweight DiT-based decoder~\cite{Peebles2022ScalableDM} $\epsilon_\psi$ then denoises the full action chunk conditioned on the per-frame point features. The initial robot state $s_t$ and diffusion step $j$ are injected via AdaLN~\cite{adaln}, and we use DDIM sampling~\cite{song2021denoising} at inference time. The full training objective $\mathcal{L}_{\mathrm{dec}}$ is provided in \cref{sec:appendix_impl}.

\section{Experimental Setup}
\label{sec:setup}
\subsection{Dataset Curation}
We curate our training corpus from BridgeData V2~\cite{bridgev2} and DROID~\cite{droid} (WidowX 250 and Franka Panda arms), filtering out samples with corrupted camera parameters or failed execution to retain $\sim$75K high-quality video clips. DROID's raw sensor depth is noisy; we recompute it from the binocular pairs with FoundationStereo~\cite{wen2025stereo}. BridgeData~V2 is monocular only, so we use Depth-Anything-V3~\cite{da3} for both metric depth and pseudo-camera intrinsics. 

\subsection{Evaluation Protocol}

\noindent\textbf{Robot Manipulation (Simulation).}
On RoboCasa365~\cite{robocasa365}, we select 10 tasks with 100 teleoperated episodes each, post-train all methods on these data, and evaluate every method in three regimes (illustrated in \cref{fig:simulation_vis}):
\begin{itemize}[leftmargin=*]
    \item \textbf{In-Distribution (ID):} same tasks and environments similar to training.
    \item \textbf{Out-of-Distribution Environments (OOD-Env):} seen tasks deployed in novel environments (novel backgrounds, varying textures), testing visual robustness.
    \item \textbf{Out-of-Distribution Tasks (OOD-Task):} novel, unseen tasks in seen environments, assessing semantic generalization and instruction following.
\end{itemize}
Each (method, regime) cell is averaged over 100 rollouts with randomized initial states.

\noindent\textbf{Cross-Embodiment Real-World.}
We additionally evaluate on two arms unseen during 4D-video pretraining: an \textbf{xArm7} and a \textbf{YAM} arm. For each arm we collect a small teleoperation set, post-train the joint RGB-XYZ model, and train the action decoder from scratch; per-task success rates are reported over independent rollouts with randomized initial object poses. Hardware, task suites, and demo budgets are detailed in \cref{sec:appendix_realworld}.

\noindent\textbf{4D Generation Quality.}
On 300 held-out DROID and BridgeData~V2 trajectories per dataset, we report PSNR/SSIM/FVD~\cite{Unterthiner2018TowardsAG} for visual quality, AbsRel and $\delta_1$ on the $z$-axis for depth, and L1 Chamfer Distance against ground truth for 3D structural fidelity.

\subsection{Baselines}
We compare against three groups of baselines (full descriptions in \cref{sec:appendix_baselines}):

\noindent\textbf{Robot Manipulation (Simulation):} recent VLA and VAM methods, namely GR00T~N1.7~\cite{gr00t}, $\pi_{0.5}$~\cite{pi05}, VPP~\cite{vpp}, and Cosmos~Policy~\cite{cosmos-policy}.

\noindent\textbf{Cross-Embodiment Real-World:} the most recent publicly available VLAs, $\pi_{0.5}$~\cite{pi05} and GR00T~N1.7~\cite{gr00t} on the xArm7 setup, and $\pi_0$~\cite{pi0} and GR00T~N1.5~\cite{gr00t} on the YAM setup; both arms use the same teleoperation, post-training, and rollout protocol described above.

\noindent\textbf{4D Generation:} state-of-the-art 4D generative models and a cascaded variant, namely TesserAct~\cite{tesseract}, 4DNeX~\cite{4dnex}, LVP~\cite{lvp}, Wan~2.1~\cite{wan}, and a cascaded \methodname~(RGB)+StreamVGGT~\cite{streamVGGT} that runs StreamVGGT on our generated RGB to obtain geometry.

%% file: sections/05_experiments.tex
\section{Results}

\subsection{Robotic Simulation Evaluation}

\input{table/table2_simulation_comparison}

\noindent\textbf{Overall performance and environmental generalization.}
\methodname attains the highest absolute success in both the ID ($47.7\%$) and OOD-Env ($44.1\%$) regimes on the 10 seen tasks (\cref{tab:simulation}), outperforming the strongest baseline Cosmos~Policy by $+2.5\%$/$+1.2\%$ respectively and the VLA baselines by larger margins ($+4.1\%$/$+6.5\%$ over GR00T~N1.7). While GR00T~N1.7 degrades severely under environment shift ($-6.0\%$), \methodname's drop ($-3.6\%$) is moderate and does not affect its top ranking in either regime.

\noindent\textbf{Zero-shot generalization to unseen tasks.}
On the 5 unseen tasks (OOD-Task row of \cref{tab:simulation}), \methodname
reaches $17.0\%$ average success, roughly $2$--$2.5\times$ the VLA baselines
(GR00T~N1.7: $8.6\%$; $\pi_{0.5}$: $6.9\%$) and above the VAM
baselines (VPP: $7.4\%$; Cosmos~Policy: $14.0\%$). Absolute success remains low, since zero-shot robotic instruction following is far from solved, but the consistent relative
margin suggests the explicit spatial interface transfers more readily to novel
task compositions than image- or latent-space conditioning.
\input{table/realworld_combined}
\subsection{Cross-Embodiment Real-World Deployment}
\label{sec:realworld}
We evaluate cross-embodiment transfer on two real arms, an \textbf{xArm7} and a \textbf{YAM} arm, both \textbf{unseen during 4D-video pretraining}, with different visual domains, task suites, and VLA baselines (full setup in \cref{sec:appendix_realworld}).

\begin{figure}[!htbp]
  \centering
  \includegraphics[width=0.95\textwidth]{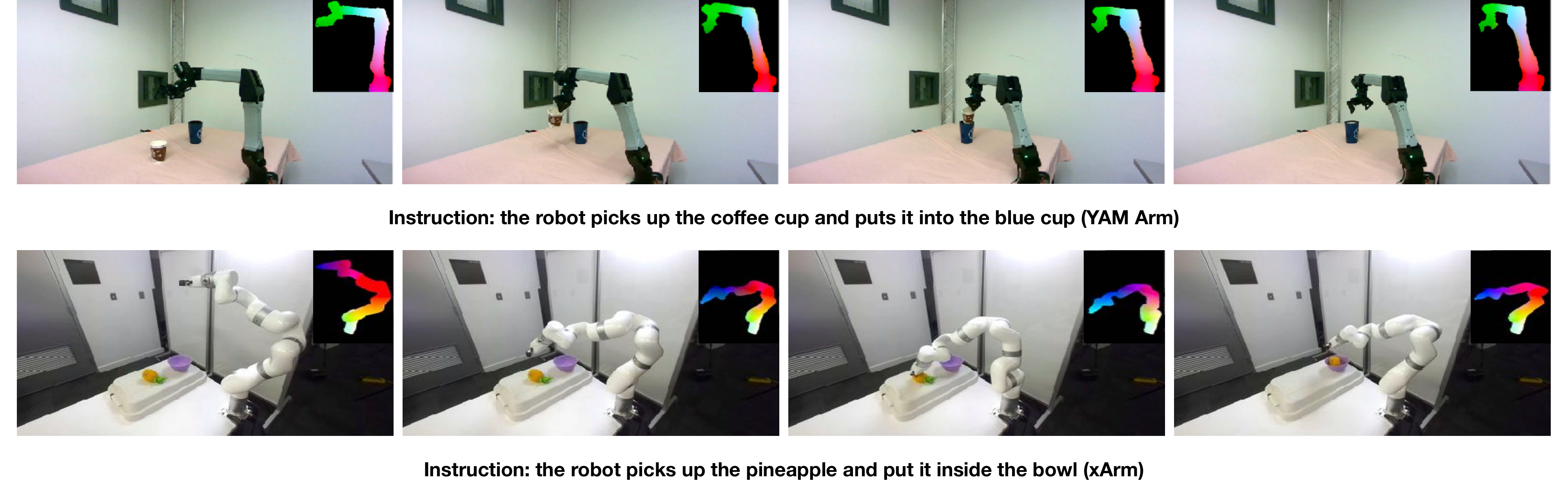}
  \caption{\small\textbf{Qualitative cross-embodiment rollouts.} \methodname executes on two real arms unseen during 4D-video pretraining: YAM (top) and xArm7 (bottom), each on a different task. Each RGB frame is annotated (top right) with the segmented robot points predicted by the 4D-VM.}
  \label{fig:realworld_qual}
\end{figure}

\methodname outperforms recent VLAs on every task and arm (\cref{tab:realworld_combined}): on xArm7 it reaches $43.0\%$ average versus $\pi_{0.5}$'s $22.7\%$, and on YAM it attains non-trivial success on tasks where baselines almost completely fail. Qualitative rollouts on both arms are shown in \cref{fig:realworld_qual}: baselines frequently hallucinate actions under novel object placements, while \methodname produces kinematically plausible trajectories even in failure cases. Consistency of these gains across two pretraining-unseen arms supports the central claim that 3D point dynamics is an embodiment-agnostic action interface.

\subsection{Ablation Study}
We jointly ablate two design choices of the action decoder---the input \emph{modality} and the \emph{point source}---in \cref{tab:ablation}.

\noindent\textbf{Modality (RGB vs.\ XYZ).} Replacing the XYZ pointmaps with RGB pixels alone causes the largest drop, from $47.7\%$ to $25.1\%$ ID success, confirming that the geometric ambiguity of pixel inputs is the dominant failure mode. Concatenating RGB on top of XYZ (Robot Only) also degrades performance ($37.2\%$ ID), suggesting that task-irrelevant visual artifacts (lighting, textures) distract the lightweight decoder and that explicit 3D geometry suffices.

\noindent\textbf{Point source (Full Scene vs.\ Robot+Scene vs.\ Robot Only).} Among XYZ-only variants, our robot-only formulation strictly dominates: feeding the full scene's points with a single point encoder performs the worst ($27.1\%$ ID), augmenting robot points with scene points (two separate encoders, latents concatenated) recovers most of the performance ($40.3\%$ ID), and masking to the robot surface gives the best results ($47.7\%$ ID, $44.1\%$ OOD-Env). Scene points carry no robot-state information and can inject ambiguity into action prediction. Substituting our jointly predicted XYZ with DepthAnything-V3~\cite{da3} depth estimated post-hoc on our generated RGB drops ID success to $28.4\%$, showing that joint RGB-XYZ generation cannot be replaced by cascaded depth estimation.
\input{table/ablation}
\subsection{4D Generation Quality}
\input{table/table1}

\cref{tab:4dgen_quantitative} verifies the 4D backbone: \methodname outperforms all baselines on both visual (RGB) and geometric (3D) metrics. Applying an off-the-shelf 4D reconstruction method~\cite{streamVGGT} to our generated 2D videos yields strictly worse 3D quality than our direct joint predictions, indicating that cascaded designs (2D RGB followed by 3D reconstruction) accumulate errors that the unified 4D generator avoids. Additional qualitative rollouts, head-to-head comparisons against 4DNeX~\cite{4dnex} / TesserAct~\cite{tesseract}, and an out-of-distribution depth study against simulator ground truth (\cref{tab:ood_recon}) are provided in \cref{sec:appendix_qual_4d}.

%% file: table/table2_simulation_comparison.tex
\begin{table}[t]
\centering
\caption{\small\textbf{Average success rates (\%) on RoboCasa365 across the three ID/OOD regimes.} Numbers are averaged over 100 rollouts per (method, task, regime) cell with randomized initial states. The full per-task breakdown is provided in the supplementary appendix. \textbf{Bold} marks the best result per row.}
\label{tab:simulation}
\resizebox{\textwidth}{!}{%
\renewcommand{\arraystretch}{1.1}%
\setlength{\tabcolsep}{10pt}%
\begin{tabular}{l ccccc}
\toprule
\textbf{Setting} & GR00T~N1.7~\cite{gr00t} & $\pi_{0.5}$~\cite{pi05} & VPP~\cite{vpp} & Cosmos~Policy~\cite{cosmos-policy} & \textbf{\methodname (Ours)} \\
\midrule
ID (10 seen tasks)        & 44.5 & 39.8 & 34.5 & 45.2 & \textbf{47.7} \\
OOD-Env (10 seen tasks)   & 37.6 & 35.2 & 32.2 & 42.9 & \textbf{44.1} \\
OOD-Task (5 unseen tasks) & \phantom{0}8.6 & \phantom{0}6.9 & \phantom{0}7.4 & 14.0 & \textbf{17.0} \\
\bottomrule
\end{tabular}}
\vspace{-8pt}
\end{table}

%% file: table/realworld_combined.tex
\begin{table}[t]
\vspace{-2pt}
\centering
\caption{\small\textbf{Cross-embodiment real-world evaluation on two arms unseen during 4D-video pretraining.} (a) xArm7: 100 rollouts per task with randomized initial object poses; all methods finetuned on 50 expert trajectories per task; compared against the most recent VLAs ($\pi_{0.5}$, GR00T~N1.7). (b) YAM arm: 20 evaluations per task; all methods finetuned on 20 expert trajectories per task; compared against $\pi_0$ and GR00T~N1.5. \methodname's point-based interface transfers cleanly to both arms, substantially outperforming VLA baselines on every task and arm.}
\label{tab:realworld_combined}
\renewcommand{\arraystretch}{1.05}
\begin{minipage}[t]{0.55\textwidth}
\centering
\textbf{(a) xArm7}\\[2pt]
{\small
\setlength{\tabcolsep}{3pt}
\renewcommand{\arraystretch}{1.05}
\begin{tabular}{l c c c c}
\toprule
Method & \makecell{Pick\\\& Place} & \makecell{Stack\\Cubes} & \makecell{Stack\\Cups} & \makecell{Avg.} \\
\midrule
GR00T~N1.7~\cite{gr00t}    & 30.0 & \phantom{0}7.0 & \phantom{0}7.0 & 14.7 \\
$\pi_{0.5}$~\cite{pi05}    & 42.0 & 12.0 & 14.0 & 22.7 \\
\textbf{Ours} & \textbf{67.0} & \textbf{28.0} & \textbf{34.0} & \textbf{43.0} \\
\bottomrule
\end{tabular}}
\end{minipage}%
\hfill
\begin{minipage}[t]{0.45\textwidth}
\centering
\textbf{(b) YAM}\\[2pt]
{\small
\setlength{\tabcolsep}{3pt}
\renewcommand{\arraystretch}{1.05}
\begin{tabular}{l c c c}
\toprule
Method & \makecell{Stack\\Cubes} & \makecell{Pick\\Pens} & \makecell{Insert\\Cups} \\
\midrule
GR00T~N1.5~\cite{gr00t}       & \phantom{0}0 & 20 & 15 \\
$\pi_0$~\cite{pi0}            & \phantom{0}0 & 10 & 15 \\
\textbf{Ours} & \textbf{20} & \textbf{60} & \textbf{50} \\
\bottomrule
\end{tabular}}
\end{minipage}
\vspace{-12pt}
\end{table}

%% file: table/ablation.tex
\begin{table}[t]
\centering
\caption{\small\textbf{Ablation on action decoder inputs.} We jointly ablate (i) the input \emph{modality} fed to the decoder (RGB pixels vs.\ XYZ pointmaps) and (ii) the \emph{point source} used for the XYZ stream (full-scene vs.\ robot+scene vs.\ robot-only). Robot-only XYZ points yield the best ID and OOD-Env performance, validating both the choice of geometric representation and the robot-centric masking. Each configuration is evaluated 100 times with randomized initial states.}
\label{tab:ablation}
\setlength{\tabcolsep}{4pt}
\renewcommand{\arraystretch}{1.05}
\begin{tabular}{l c c}
\toprule
Decoder Input & Success Rates (ID) $\uparrow$ & Success Rates (OOD-Env) $\uparrow$ \\
\midrule
RGB Only                          & 25.1 & 20.3 \\
RGB + XYZ (Robot Only)            & 37.2 & 30.9 \\
\midrule
XYZ Only -- Full Scene            & 27.1 & 19.4 \\
XYZ Only -- Robot + Scene         & 40.3 & 33.7 \\
XYZ Only -- Robot Only (DA3 source) & 28.4 & 21.7 \\
\textbf{XYZ Only -- Robot Only (Ours)} & \textbf{47.7} & \textbf{44.1} \\
\bottomrule
\end{tabular}
\vspace{-8pt}
\end{table}

%% file: table/table1.tex
\newcommand{\tss}{\textsuperscript}
\newcommand{\tdag}{\textdagger}

\setlength{\tabcolsep}{6pt}

\begin{table}[tb]
  \centering
  \caption{\small
  \textbf{Results on 4D Robot Scene Generation.} We select 300 unseen samples from DROID\cite{droid} and BridgeData V2\cite{bridgev2} and report average metrics on the generated results. Best per metric in \textbf{bold}. \methodname achieves best across all metrics.
  }
  \scalebox{0.9}{
  \begin{tabular}{@{}l|ccc|cc|c@{}}
    \toprule
    & \multicolumn{3}{c}{RGB} 
    & \multicolumn{2}{c}{Depth} 
    & \multicolumn{1}{c}{Point Cloud} \\
    \cmidrule(lr){2-4} \cmidrule(lr){5-6} \cmidrule(lr){7-7}
    Methods 
    & PSNR $\uparrow$ 
    & SSIM $\uparrow$ 
    & FVD $\downarrow$ 
    & AbsRel $\downarrow$ 
    & $\delta_1$ $\uparrow$ 
    & Chamfer $L_1$ $\downarrow$ \\
    \midrule
    \midrule
    {TesserAct\cite{tesseract}} & 12.225 & 0.487 & 746 & 0.403 & 0.641 & 0.389\\ %
    {4DNeX\cite{4dnex}} & 13.858 & 0.542 & 818 & 0.348 & 0.681 & 0.370 \\ %
    LVP\cite{lvp}  & 19.613 & 0.816 & 330 & - & - & - \\ %
    Wan 2.1 14B \cite{wan}& 14.532 & 0.674 & 671 & - & - & -\\%
    \midrule
    Ours* (w. StreamVGGT\cite{streamVGGT}) & - & - & - & 0.382 & 0.675 & 0.341\\
    Ours & \textbf{19.631} & \textbf{0.821} & \textbf{320} & 
    \textbf{0.176} & \textbf{0.890} & \textbf{0.122} \\ 

  \bottomrule
  \end{tabular}
  }
  
  \label{tab:4dgen_quantitative}
\vspace{-12pt}
\end{table}

%% file: sections/06_disucssion.tex
\section{Conclusion and Limitation}
In this work, we introduced \methodname, a novel framework that seamlessly bridges the gap between generative video priors and robot control by grounding both into a point-based universal action space. By formulating 4D world modeling as the joint generation of RGB frames and XYZ pointmaps, \methodname successfully relieves the geometric ambiguity inherent to 2D video-action models, providing a scalable paradigm for learning action representations through a 3D interface. 

However, \methodname is currently bottlenecked by the inherently slow inference of video models and relies on open-loop execution via a single forward pass, making it vulnerable to compounding errors under physical disturbances. A promising future direction is distilling the video backbone into an efficient autoregressive architecture~\cite{huang2025self, causvid, zhu2026causal}. Coupled with KV caching, this would enable real-time, closed-loop control, effectively mitigating both inference latency and execution errors.

%% file: sections/07_appendix.tex
% =============================================================================
% Appendix root file. Inputs from sections/appendix/.
% Main.tex switches to appendix numbering before \input{sections/07_appendix}.
% Appendix sections mirror the order of the main paper's Method + Results.
% =============================================================================

\section*{Appendix Overview}
This appendix complements the main paper as follows.
\cref{sec:appendix_impl} provides extended implementation and training details for the joint RGB-XYZ video model and the action decoder, including hyperparameters and per-task baseline configurations.
\cref{sec:appendix_sim_tasks} expands the simulation evaluation (\cref{tab:simulation}) with an illustration of the three ID/OOD regimes, qualitative dynamics--action alignment plots, and per-task rollout visualizations across all 15 RoboCasa365 tasks.
\cref{sec:appendix_realworld} details the cross-embodiment real-world setup of \cref{sec:realworld}: hardware, task suites, and demonstration budgets for both the xArm7 and the YAM arm, together with qualitative YAM rollouts.
\cref{sec:appendix_qual_4d} provides additional analyses of the 4D generation backbone: qualitative rollouts illustrating temporal and geometric consistency, head-to-head comparisons with 4D-generation baselines, and an out-of-distribution depth-reconstruction study against simulator ground truth.
\cref{sec:appendix_failure} discusses the dominant failure modes we observe in deployment.
Real-world execution videos for the YAM-arm tasks are included in the supplementary ZIP under \texttt{videos/}.

\input{sections/appendix/A_implementation}

\input{sections/appendix/B_simulation_tasks}
\input{sections/appendix/C_realworld}

\input{sections/appendix/D_qualitative_4d}

\input{sections/appendix/E_failure_modes}

%% file: sections/appendix/A_implementation.tex
\section{Extended Implementation Details}
\label{sec:appendix_impl}

\subsection{Model Architecture}
Our joint RGB-XYZ model retains the original LVP~\cite{lvp} image-to-video (I2V) backbone---comprising the text encoder, video VAE, and transformer---while extending it to concurrently model RGB frames and XYZ pointmaps. Both RGB and XYZ videos are encoded with the shared VAE. The resulting XYZ latents are normalized to a standard normal distribution using pre-computed dataset statistics. The two latent tensors are then concatenated along the width dimension, effectively doubling the spatial token width and the maximum token budget. To enable the shared transformer to disambiguate between modalities, we introduce a lightweight modality embedding using separate learnable vectors for RGB and XYZ. We additionally repeat the Rotary Position Embedding (RoPE) along the width dimension to accommodate the doubled layout. The decoder separates the output back into distinct RGB and XYZ streams. The model is trained using the standard latent flow-matching objective. During LoRA~\cite{lora} fine-tuning, we update approximately 209M parameters while keeping the text encoder and VAE frozen.

For the DiT-based action decoder, each frame's robot-centric point cloud is encoded into a per-frame feature vector using a shared, 3-layer PointNet-style~\cite{Qi2016PointNetDL} MLP. The diffusion model operates directly on the full 49-step action sequence. We concatenate the noisy action tokens with the per-frame point features, project them to a DiT hidden dimension of 256, add learnable positional embeddings, and process them through a stack of AdaLN-modulated Transformer blocks. Global conditioning---comprising the diffusion timestep and the first-frame robot state---is injected exclusively via AdaLN layers to cleanly separate modality roles. The prediction head outputs the added noise ($\epsilon$) following a standard 100-timestep diffusion schedule. During inference, we employ a DDIM-style~\cite{song2021denoising} sampler for 10 steps, allowing the decoder to predict the full action sequence in a single shot.

\subsection{4D Generation Training Details}
The joint RGB-XYZ generation model is trained on a combined dataset of 75K trajectories: 50K from DROID~\cite{droid} and 25K from BridgeData~V2~\cite{bridgev2}. All training videos are uniformly downsampled to 49 frames and spatially resized to $832\times 480$. Detailed training hyperparameters are summarized in \cref{tab:hyperparams}. During inference, we utilize a UniPC sampler with 40 denoising steps and a default text classifier-free guidance (CFG) scale of 2.5. A single inference pass takes approximately 6 minutes on a single NVIDIA B200 GPU. Optionally, history (first-frame) conditioning can be enabled to yield higher generation quality at the cost of increased sampling time.

\paragraph{Flow-matching objective.}
Each sequence of joint latents $z^{joint}$ is randomly split into a history context $\hat{z}$ (of length $m$) and a future trajectory $\tilde{z}$. Independent noise levels are applied to each. For the future frames, given a noise level $\tau \in [0, 1]$ and standard Gaussian noise $\epsilon \sim \mathcal{N}(0, I)$, the noisy target is constructed via linear interpolation:
\begin{equation}
\tilde{z}_{\tau} = (1-\tau)\,\tilde{z} + \tau\,\epsilon.
\end{equation}
The history $\hat{z}$ is diffused with an independent noise level $\tau'$, which we set to $\tau' = 0$ (clean context) with $50\%$ probability to encourage robust conditioning. The joint transformer backbone $v_\theta$ is then optimized to predict the flow field $v = (\epsilon - \tilde{z})$, conditioned on $l$:
\begin{equation}
\mathcal{L}_{\mathrm{flow}} = \mathbb{E}_{z^{joint}, \epsilon, \tau, \tau', m}\left[\, \big\| v_\theta(\tilde{z}_{\tau}, \hat{z}_{\tau'}, l, \tau) - v \big\|_2^2\, \right].
\end{equation}
\input{table/supp_vid_config}

\subsection{Action Decoder Training (Simulation)}
For evaluation on RoboCasa365, we additionally post-train the 4D model for 1{,}000 iterations on each RoboCasa task split and train the DiT-based action decoder from scratch. Per-task data, baseline implementations, and training configurations are summarized below.

\paragraph{Decoder objective.}
Per-frame robot-centric points $\tilde u_{\mathrm{robo}}$ are downsampled by FPS to $\mathcal{\tilde{P}} = \mathrm{FPS}(\tilde{u}_{\mathrm{robo}}, N)$ with $N{=}512$ and encoded by a PointNet-style MLP $\Phi$. Let $a$ denote the ground-truth action sequence. At diffusion step $j$ we inject Gaussian noise $\epsilon' \sim \mathcal{N}(0, I)$ to obtain a noisy action $a^{(j)}$, and train the DiT-based decoder $\epsilon_\psi$ via the standard $\epsilon$-prediction loss conditioned on the extracted point features and the initial robot state $s_t$:
\begin{equation}
\mathcal{L}_{\mathrm{dec}} = \mathbb{E}_{a, \epsilon', j}\left[\, \big\| \epsilon' - \epsilon_\psi\!\big(a^{(j)}, \Phi(\mathcal{\tilde{P}}), s_t, j\big) \big\|_2^2\, \right].
\end{equation}

\paragraph{Datasets.} Each simulation task comprises approximately 100 expert episodes collected via human teleoperation. For \methodname, trajectories are uniformly downsampled to 49 frames prior to training.

\paragraph{Baseline Implementations.}
For $\pi_0$~\cite{pi0} and Isaac GR00T~N1.5~\cite{gr00t}, we use the implementations provided by the RoboCasa365~\cite{robocasa365} benchmark and initialize from their official \texttt{pi0\_base} and \texttt{nvidia/GR00T-N1.5-3B} checkpoints. For Cosmos~Policy~\cite{cosmos-policy} and VPP~\cite{vpp}, we use the official codebases. Cosmos~Policy is fine-tuned from \texttt{nvidia/Cosmos-Policy-RoboCasa-Predict2-2B}. For VPP, the Stable Video Diffusion backbone is fine-tuned from its public pre-trained weights, while the action-decoding head is trained from scratch. All baselines are fine-tuned on the same RoboCasa data split as \methodname.

\paragraph{Training Configurations.} The DiT-based action decoder uses 6 transformer blocks with hidden dimension 256 and 4 attention heads. We train for 3{,}000 epochs with batch size 32 using AdamW (learning rate $5\times10^{-4}$, weight decay $1\times10^{-4}$) under a cosine learning-rate schedule with 1{,}000 warmup steps. For point cloud augmentation, we apply random point dropout (up to 20\%) and isotropic Gaussian jitter ($\sigma{=}0.05$).

\subsection{Baseline Details}
\label{sec:appendix_baselines}
This section expands the compact baseline list in the main paper (\cref{sec:setup}).

\paragraph{Robot Manipulation (Simulation).}
\begin{itemize}
    \item \textbf{GR00T~N1.5}~\cite{gr00t}: a VLA combining a flow-matching diffusion transformer with synthetic-data integration for cross-embodiment robotic control.
    \item $\bm{\pi_0}$~\cite{pi0}: a general-purpose VLA combining a pretrained VLM backbone~\cite{paligemma} with a flow-matching action expert for continuous control across embodiments.
    \item \textbf{VPP}~\cite{vpp}: a decoupled Video-Action framework built on Stable Video Diffusion (SVD)~\cite{svd}; the SVD backbone is pretrained on robot and human video, and an inverse dynamics model decodes actions from the SVD latent.
    \item \textbf{Cosmos~Policy}~\cite{cosmos-policy}: built on Cosmos-Predict2~\cite{cosmos-predict2}, jointly generating video and action encoded as latent frames within the video diffusion process.
\end{itemize}

\paragraph{Cross-Embodiment Real-World.}
On the xArm7 setup we upgrade the VLA baselines to the most recent publicly available versions, $\bm{\pi_{0.5}}$~\cite{pi05} and \textbf{GR00T~N1.7}~\cite{gr00t}; on the YAM setup we use $\bm{\pi_0}$ and \textbf{GR00T~N1.5}. All baselines are finetuned on the same per-arm teleoperation set used by \methodname{} and evaluated under the same rollout budget (see \cref{sec:appendix_realworld}).

\paragraph{4D Generation.}
\begin{itemize}
    \item \textbf{TesserAct}~\cite{tesseract}: predicts RGB-DN (RGB, depth, and normal) sequences for robotic scenes, refined by normal integration into a 4D representation.
    \item \textbf{4DNeX}~\cite{4dnex}: a feed-forward 4D generation model built on the Wan~2.1 architecture, trained on large-scale 4D datasets to jointly synthesize RGB and XYZ.
    \item \textbf{Large Video Planner (LVP)}~\cite{lvp}: a foundation video model tailored for robotic tasks and trained across human and robot interaction video.
    \item \textbf{Wan~2.1 (14B)}~\cite{wan}: a large-scale pretrained video diffusion model for zero-shot generation across visual domains.
    \item \textbf{\methodname (RGB) + StreamVGGT}~\cite{streamVGGT}: a cascaded variant in which our model's generated 2D RGB videos are passed through StreamVGGT, a state-of-the-art dynamic reconstruction model, and the extracted geometry is compared against our direct joint XYZ predictions.
\end{itemize}

%% file: table/supp_vid_config.tex
\begin{table}[htbp]
\centering
\caption{\textbf{Hyperparameters for the RGB-XYZ Model.}}
\label{tab:hyperparams}
\resizebox{0.6\textwidth}{!}{
\begin{tabular}{lc}
\toprule
Hyperparameter & Value \\
\midrule
Optimizer & AdamW \\
Learning Rate & $1 \times 10^{-4}$ \\
Weight Decay & $5 \times 10^{-2}$ \\
$\beta_1, \beta_2$ & $0.9, 0.95$ \\
Global Batch Size & $16$ \\
\midrule
LoRA Rank / $\alpha$ / Dropout & $128$ / $64$ / $0.0$ \\
XYZ Latent Norm ($\mu, \sigma$) & ($-0.227444$, $1.437663$) \\
\bottomrule
\end{tabular}
}
\end{table}

%% file: sections/appendix/B_simulation_tasks.tex
\section{Detailed Simulation Results}
\label{sec:appendix_sim_tasks}

This appendix provides the full per-task numerical breakdown that supports the averages in \cref{tab:simulation} of the main paper, together with visualizations of the evaluation regimes and per-task rollouts.

\subsection{Evaluation Settings}
\begin{figure}[!htbp]
    \centering
    \includegraphics[width=0.85\textwidth]{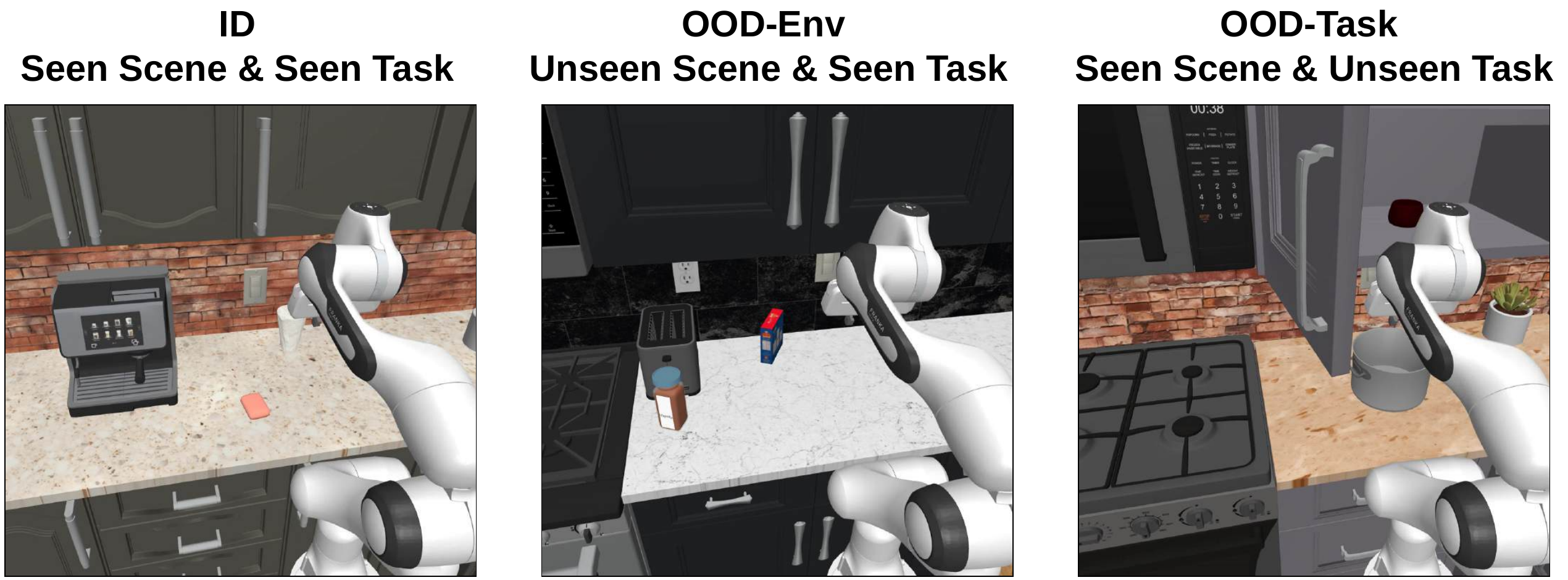}
    \caption{\textbf{Visualization of simulation evaluation settings.}}
    \label{fig:simulation_vis}
\end{figure}

\cref{fig:simulation_vis} illustrates the three evaluation regimes used in \cref{tab:simulation,tab:simulation_per_task}: \textbf{ID} (seen scene and seen task), \textbf{OOD-Env} (unseen scene, seen task), and \textbf{OOD-Task} (seen scene, unseen task).

\subsection{Per-Task Success Rates}
\cref{tab:simulation_per_task} expands \cref{tab:simulation} into per-task success rates across all 15 RoboCasa365 tasks (10 seen, 5 unseen). The top block reports both ID and OOD-Env success per seen task, while the bottom block reports OOD-Task success on five completely unseen tasks.
\input{table/simulation_per_task}

\subsection{Dynamics-Action Alignment}
\begin{figure}[!htbp]
    \centering
    \includegraphics[width=0.95\textwidth]{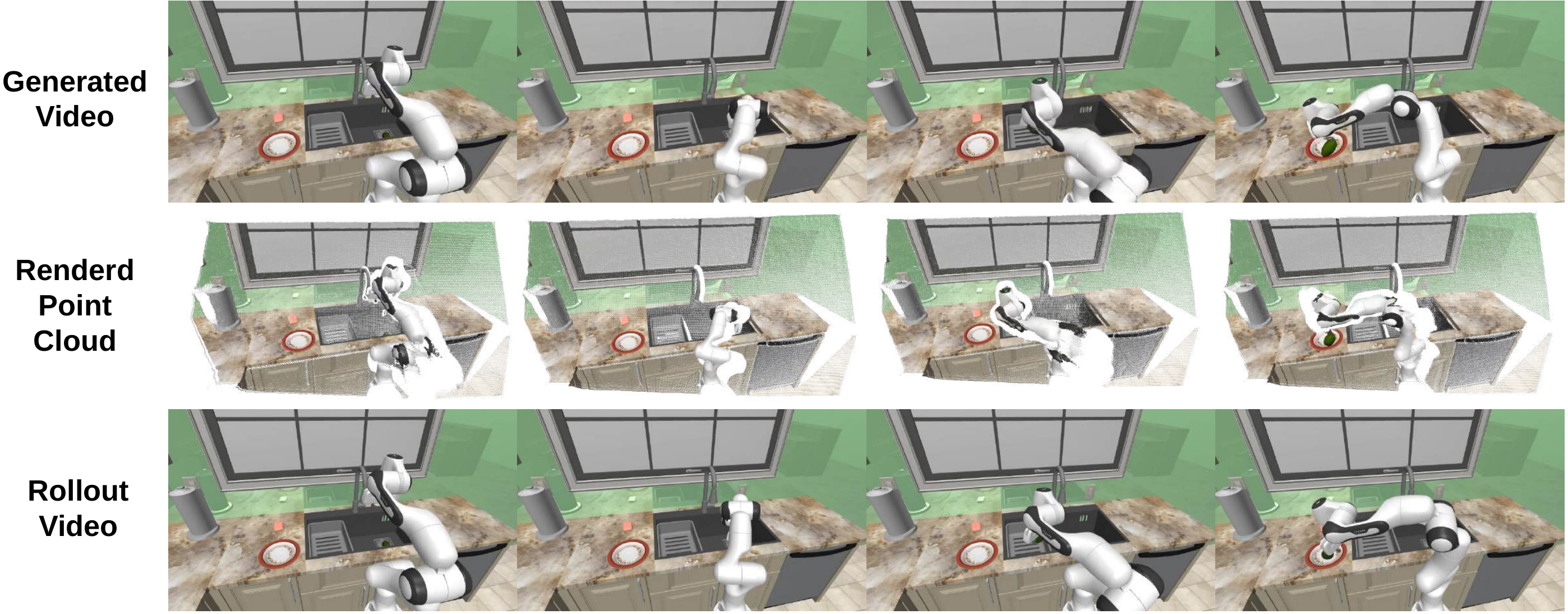}
    \caption{\textbf{Qualitative results on RoboCasa365.} Predicted video frames (top) and actual physical robot execution (bottom) for each task.}
    \label{fig:simulation_qual_result}
\end{figure}

\cref{fig:simulation_qual_result} qualitatively compares the joint RGB-XYZ prediction with the actual physical robot execution. The generated videos and the actual rollouts align spatio-temporally, confirming that the predicted point dynamics are faithful to the resulting motion.

\subsection{Per-Task Rollouts}
We visualize one representative rollout per RoboCasa365 task used in our simulation evaluation. Each row shows a sequence of sampled keyframes from left to right.

\begin{figure}[!htbp]
    \centering
    \includegraphics[width=\textwidth, height=0.10\textheight, keepaspectratio]{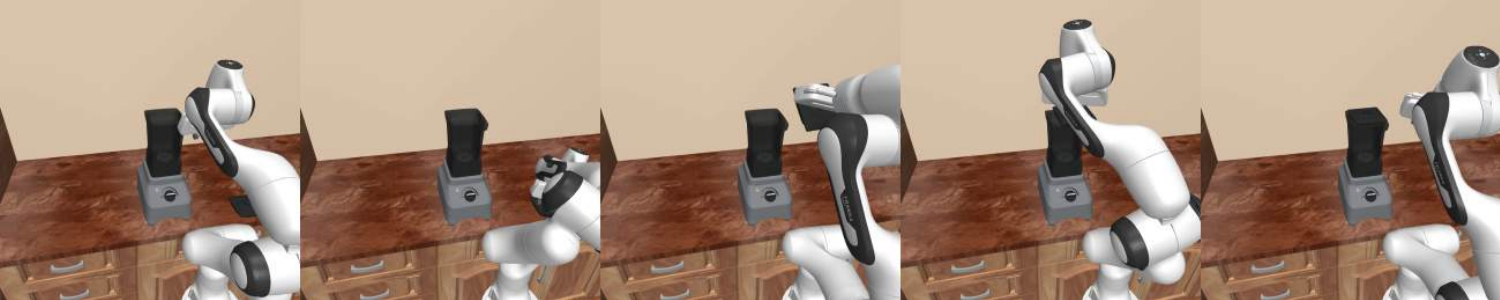}
    \caption{\textbf{CloseBlenderLid}}
    \label{fig:sim_first}
\end{figure}
\begin{figure}[!htbp]
    \centering
    \includegraphics[width=\textwidth, height=0.10\textheight, keepaspectratio]{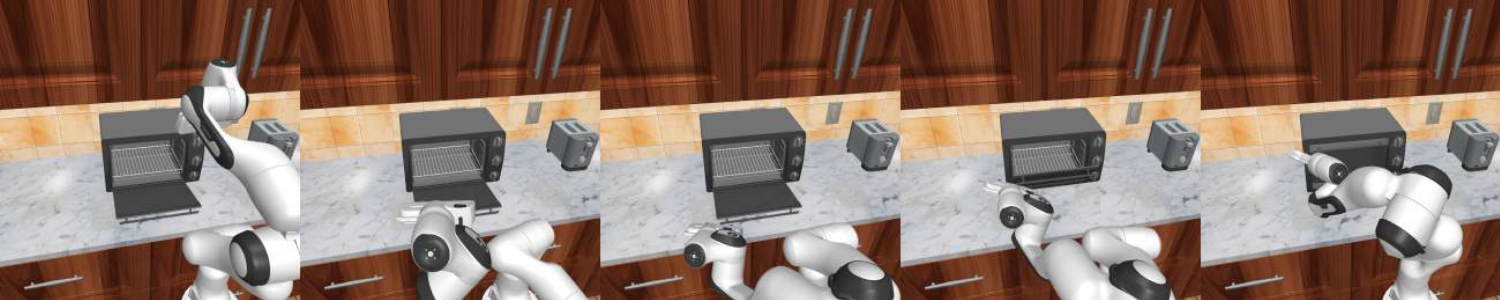}
    \caption{\textbf{CloseToasterOvenDoor}}
\end{figure}
\begin{figure}[!htbp]
    \centering
    \includegraphics[width=\textwidth, height=0.10\textheight, keepaspectratio]{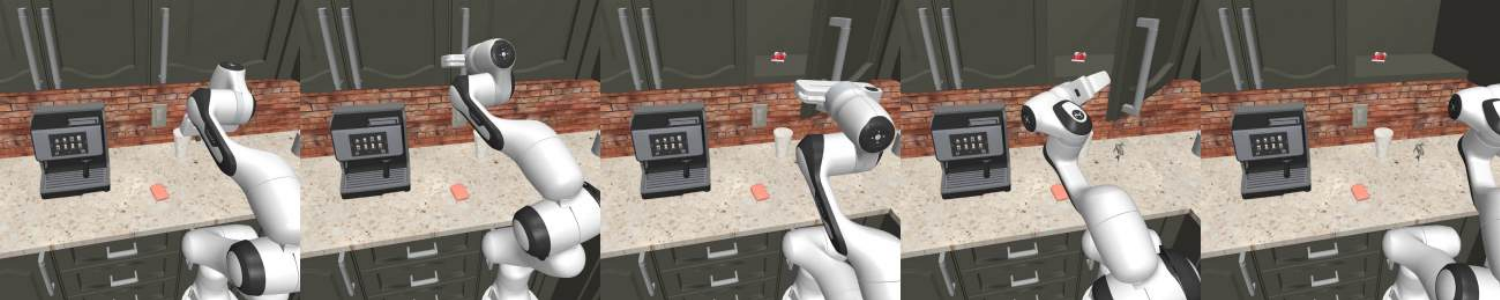}
    \caption{\textbf{OpenCabinet}}
\end{figure}
\begin{figure}[!htbp]
    \centering
    \includegraphics[width=\textwidth, height=0.10\textheight, keepaspectratio]{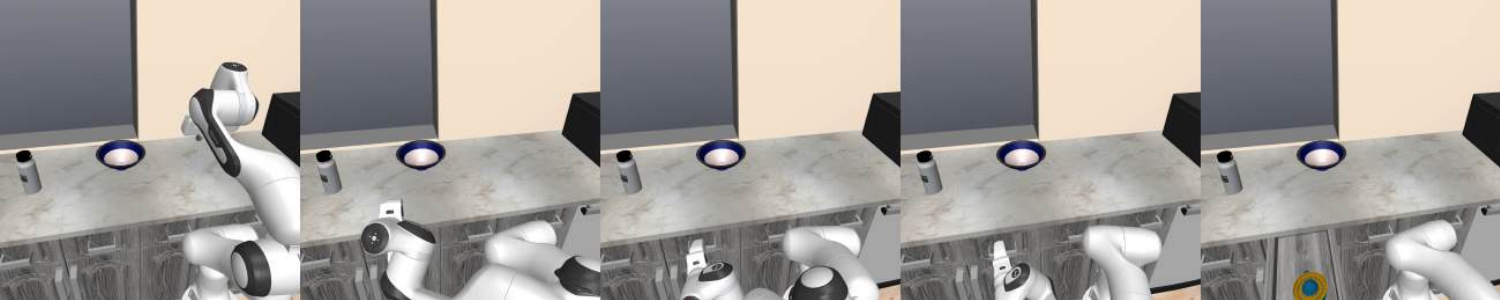}
    \caption{\textbf{OpenDrawer}}
\end{figure}
\begin{figure}[!htbp]
    \centering
    \includegraphics[width=\textwidth, height=0.10\textheight, keepaspectratio]{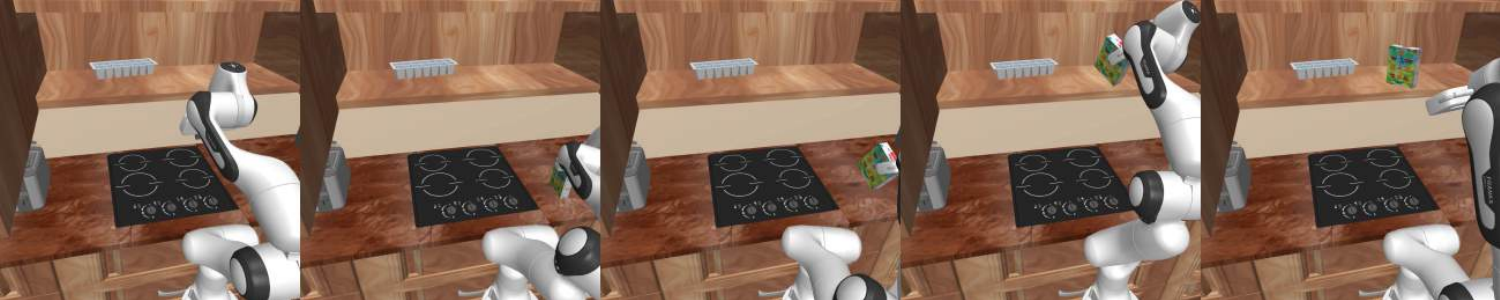}
    \caption{\textbf{PickPlaceCounterToCabinet}}
\end{figure}
\begin{figure}[!htbp]
    \centering
    \includegraphics[width=\textwidth, height=0.10\textheight, keepaspectratio]{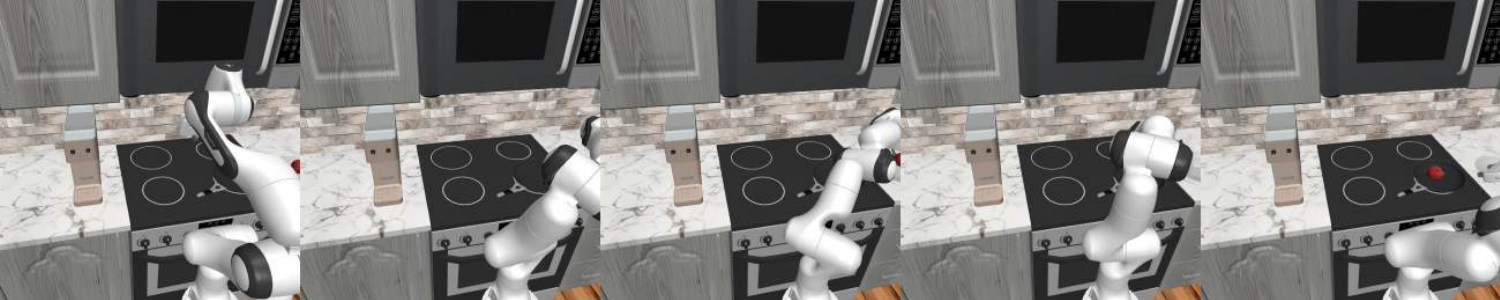}
    \caption{\textbf{PickPlaceCounterToStove}}
\end{figure}
\begin{figure}[!htbp]
    \centering
    \includegraphics[width=\textwidth, height=0.10\textheight, keepaspectratio]{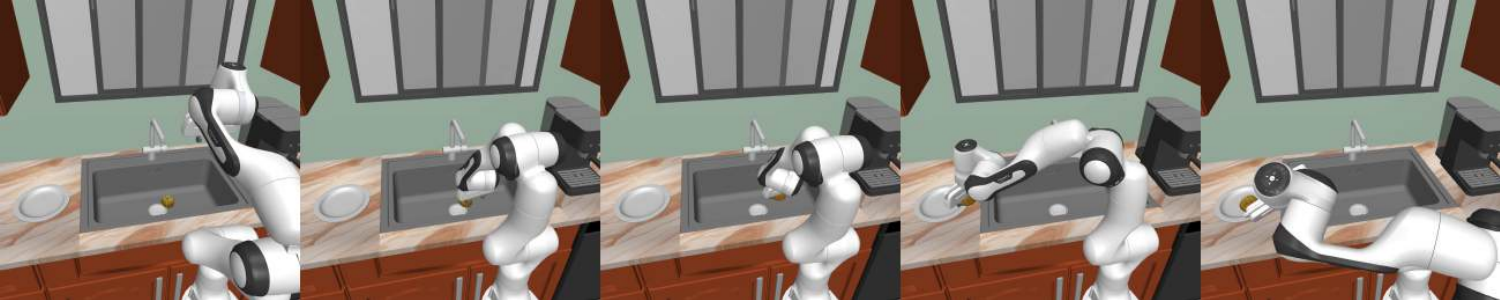}
    \caption{\textbf{PickPlaceSinkToCounter}}
\end{figure}
\begin{figure}[!htbp]
    \centering
    \includegraphics[width=\textwidth, height=0.10\textheight, keepaspectratio]{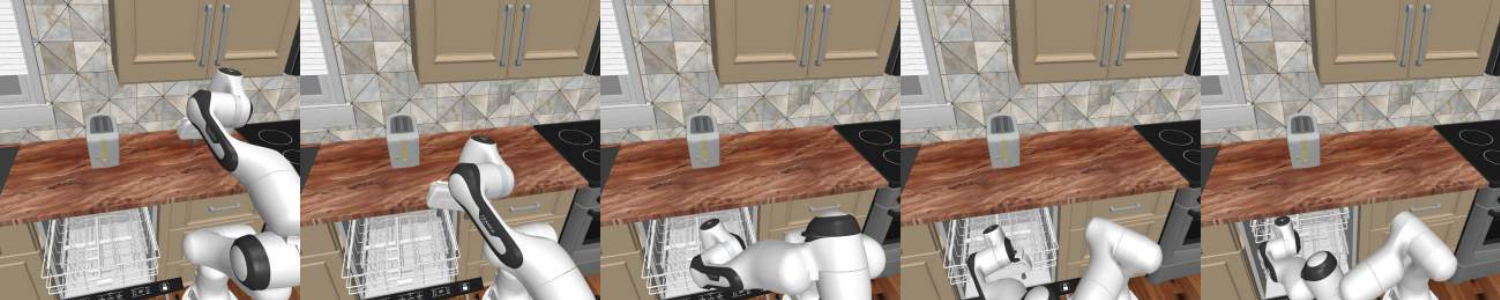}
    \caption{\textbf{SlideDishwasherRack}}
\end{figure}
\begin{figure}[!htbp]
    \centering
    \includegraphics[width=\textwidth, height=0.10\textheight, keepaspectratio]{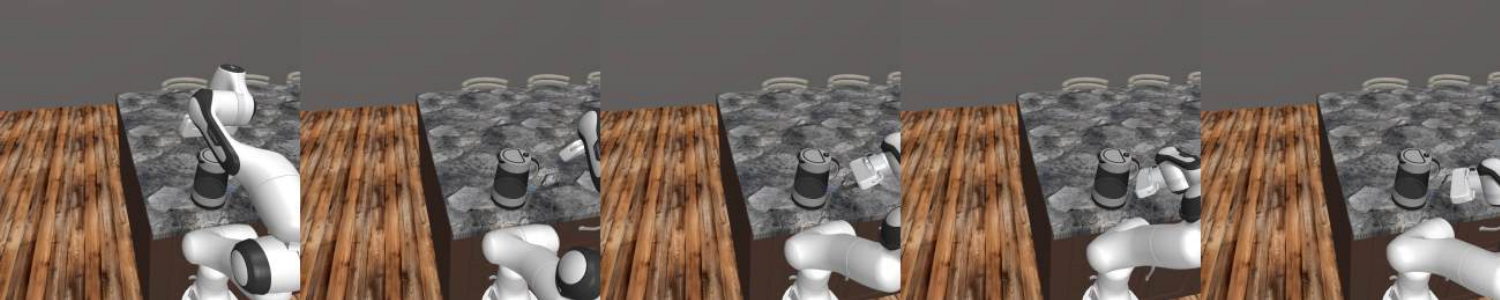}
    \caption{\textbf{TurnOnElectricKettle}}
\end{figure}
\begin{figure}[!htbp]
    \centering
    \includegraphics[width=\textwidth, height=0.10\textheight, keepaspectratio]{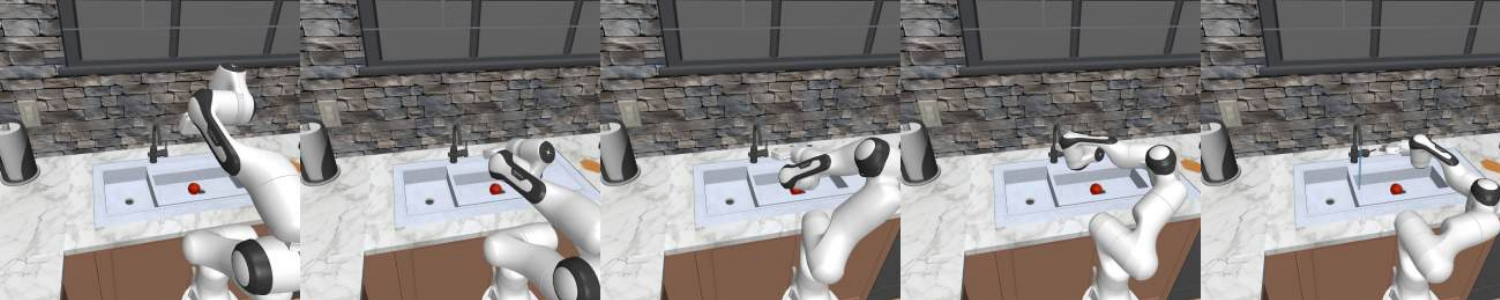}
    \caption{\textbf{TurnOnSinkFaucet}}
\end{figure}
\begin{figure}[!htbp]
    \centering
    \includegraphics[width=\textwidth, height=0.10\textheight, keepaspectratio]{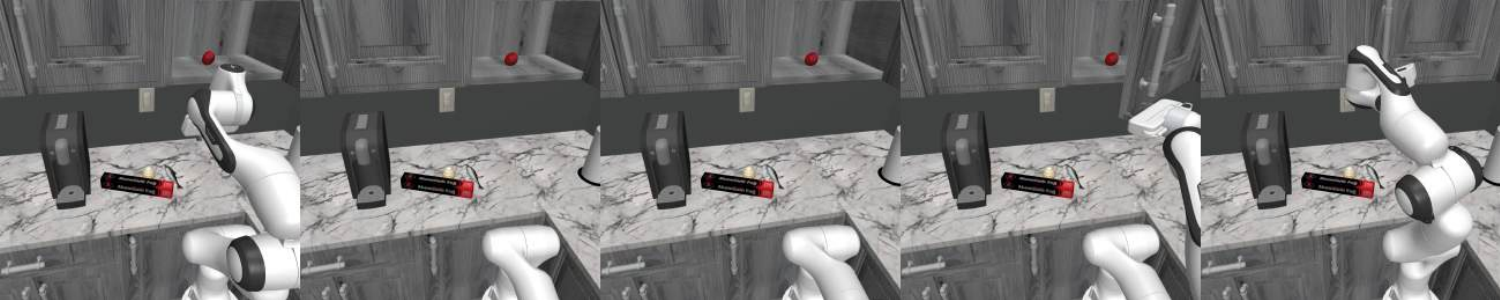}
    \caption{\textbf{CloseCabinet}}
\end{figure}
\begin{figure}[!htbp]
    \centering
    \includegraphics[width=\textwidth, height=0.10\textheight, keepaspectratio]{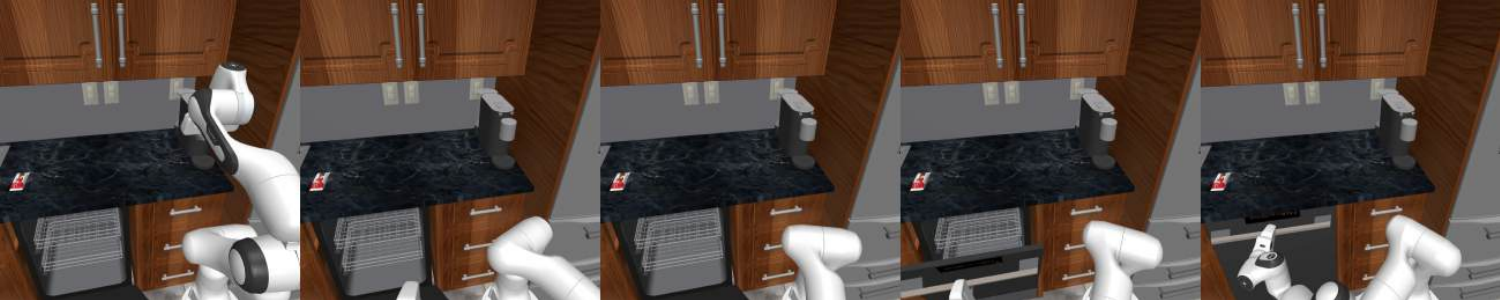}
    \caption{\textbf{CloseDishwasher}}
\end{figure}
\begin{figure}[!htbp]
    \centering
    \includegraphics[width=\textwidth, height=0.10\textheight, keepaspectratio]{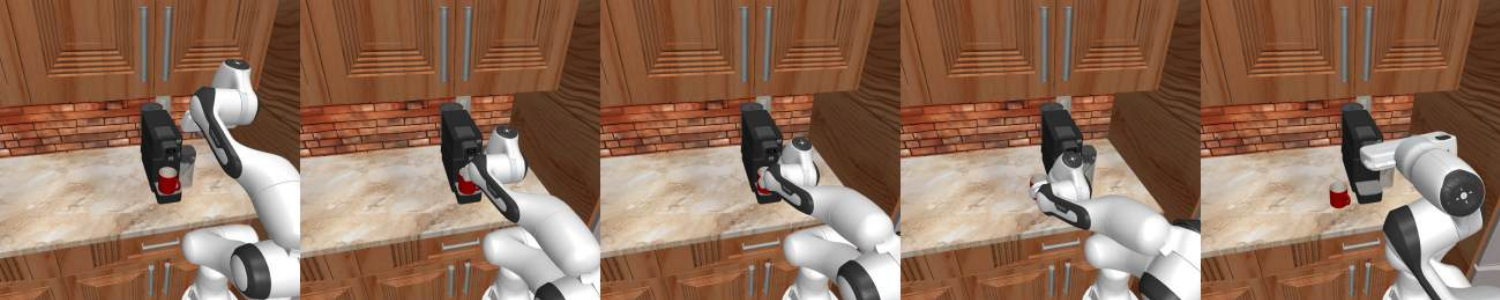}
    \caption{\textbf{CoffeeServeMug}}
\end{figure}
\begin{figure}[!htbp]
    \centering
    \includegraphics[width=\textwidth, height=0.10\textheight, keepaspectratio]{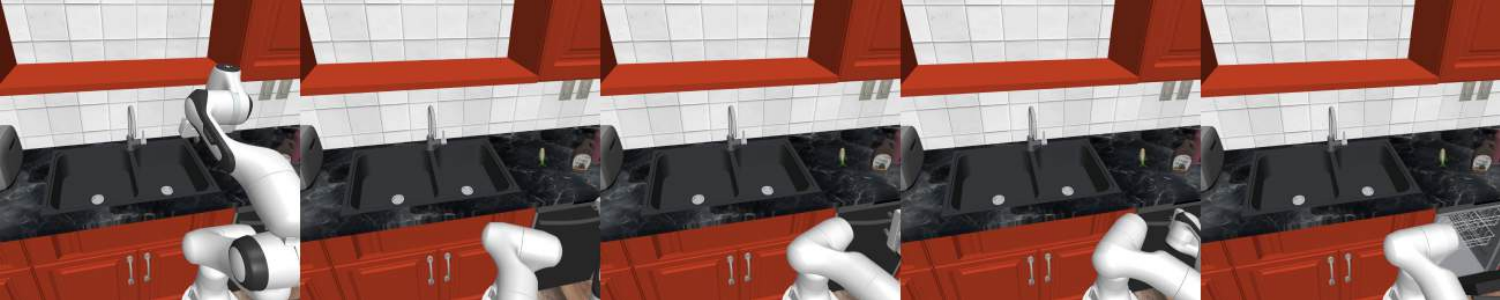}
    \caption{\textbf{OpenDishwasher}}
\end{figure}
\begin{figure}[!htbp]
    \centering
    \includegraphics[width=\textwidth, height=0.10\textheight, keepaspectratio]{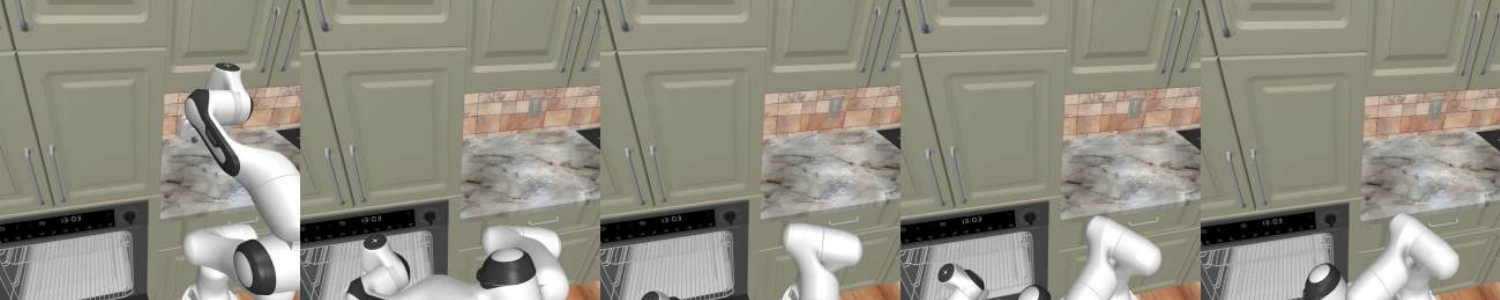}
    \caption{\textbf{SlideOvenRack}}
    \label{fig:sim_last}
\end{figure}

%% file: table/simulation_per_task.tex
\begin{table}[!htbp]
\centering
\caption{\textbf{Per-task success rates (\%) on RoboCasa365 across the three ID/OOD regimes.} \emph{Top block:} 10 training tasks evaluated in In-Distribution (ID) and Out-of-Distribution Environment (OOD-Env) regimes; each cell reports ID\,$\mid$\,OOD-Env. \emph{Bottom block:} 5 completely unseen tasks evaluated in the OOD-Task regime (zero-shot instruction following); each cell reports a single OOD-Task success rate. Every (method, task, regime) cell is computed from 100 rollouts with randomized initial state. \textbf{Bold} marks the best result per row.}
\label{tab:simulation_per_task}
\resizebox{\textwidth}{!}{
\begin{tabular}{l ccccc}
\toprule
\textbf{Task} & GR00T~N1.5~\cite{gr00t} & $\pi_0$~\cite{pi0} & VPP~\cite{vpp} & Cosmos~Policy~\cite{cosmos-policy} & \textbf{\methodname (Ours)} \\
\midrule
\multicolumn{6}{l}{\textit{Seen tasks --- ID\,$\mid$\,OOD-Env}} \\
CloseBlenderLid             & \phantom{0}4 $\mid$ \phantom{0}1 & \phantom{0}4 $\mid$ \phantom{0}4 & \phantom{0}3 $\mid$ \phantom{0}2 & \textbf{\phantom{0}5} $\mid$ \textbf{\phantom{0}5} & \phantom{0}1 $\mid$ \phantom{0}0 \\
CloseToasterOvenDoor        & 55 $\mid$ 45 & \textbf{73} $\mid$ 58 & 54 $\mid$ 48 & 59 $\mid$ 57 & 67 $\mid$ \textbf{63} \\
OpenCabinet                 & 33 $\mid$ 30 & 31 $\mid$ 24 & 29 $\mid$ 28 & 35 $\mid$ 34 & \textbf{42} $\mid$ \textbf{39} \\
OpenDrawer                  & 44 $\mid$ 40 & 25 $\mid$ 22 & 27 $\mid$ 29 & 57 $\mid$ 59 & \textbf{70} $\mid$ \textbf{66} \\
PickPlaceCounterToCabinet   & 39 $\mid$ 31 & 25 $\mid$ 24 & 29 $\mid$ 26 & 40 $\mid$ 36 & \textbf{46} $\mid$ \textbf{42} \\
PickPlaceCounterToStove     & 32 $\mid$ 34 & \phantom{0}9 $\mid$ 10 & 19 $\mid$ 17 & \textbf{37} $\mid$ \textbf{35} & 31 $\mid$ 25 \\
PickPlaceSinkToCounter      & \textbf{59} $\mid$ \textbf{62} & 31 $\mid$ 24 & 30 $\mid$ 27 & 48 $\mid$ 44 & 39 $\mid$ 36 \\
SlideDishwasherRack         & 69 $\mid$ 66 & \textbf{79} $\mid$ \textbf{71} & 62 $\mid$ 61 & 63 $\mid$ 59 & 66 $\mid$ 63 \\
TurnOnElectricKettle        & 57 $\mid$ 54 & 51 $\mid$ 43 & 53 $\mid$ 50 & 62 $\mid$ 57 & \textbf{68} $\mid$ \textbf{63} \\
TurnOnSinkFaucet            & 44 $\mid$ 26 & 45 $\mid$ 41 & 39 $\mid$ 34 & 46 $\mid$ 43 & \textbf{47} $\mid$ \textbf{44} \\
\cmidrule(lr){1-6}
\textbf{Seen-task Avg.}     & 43.6 $\mid$ 38.9 & 37.3 $\mid$ 32.1 & 34.5 $\mid$ 32.2 & 45.2 $\mid$ 42.9 & \textbf{47.7} $\mid$ \textbf{44.1} \\
\midrule
\multicolumn{6}{l}{\textit{Unseen tasks --- OOD-Task (zero-shot)}} \\
CloseCabinet           & 0  & 0  & 5  & 13  & \textbf{17} \\
CloseDishwasher        & 0  & 0  & 8  & 11  & \textbf{14} \\
CoffeeServeMug         & 1  & 1  & 6  & \textbf{9}  & 2 \\
OpenDishwasher         & 0  & 5  & 4  & 18  & \textbf{25} \\
SlideOvenRack          & 9  & 17 & 14 & 19  & \textbf{27} \\
\cmidrule(lr){1-6}
\textbf{Unseen-task Avg.} & 2.0 & 4.6 & 7.4 & 14.0 & \textbf{17.0} \\
\bottomrule
\end{tabular}
}
\end{table}

%% file: sections/appendix/C_realworld.tex
\section{Cross-Embodiment Real-World Details}
\label{sec:appendix_realworld}

\cref{sec:realworld} reports cross-embodiment results on two real robotic arms---xArm7 and YAM. Both arms were \textbf{unseen during 4D-video pretraining}; they differ in hardware, visual domain, task suite, and the VLA baselines they are compared against. We detail the per-arm setup, tasks, and training protocol below. The corresponding combined success-rate table is \cref{tab:realworld_combined} and qualitative rollouts on both arms are visualized in \cref{fig:realworld_qual} of the main paper. Execution videos for both arms are included in the supplementary materials.

\subsection{xArm7 Setup}
\label{sec:appendix_xarm}
\begin{figure}[!htbp]
  \centering
  \includegraphics[width=0.75\textwidth]{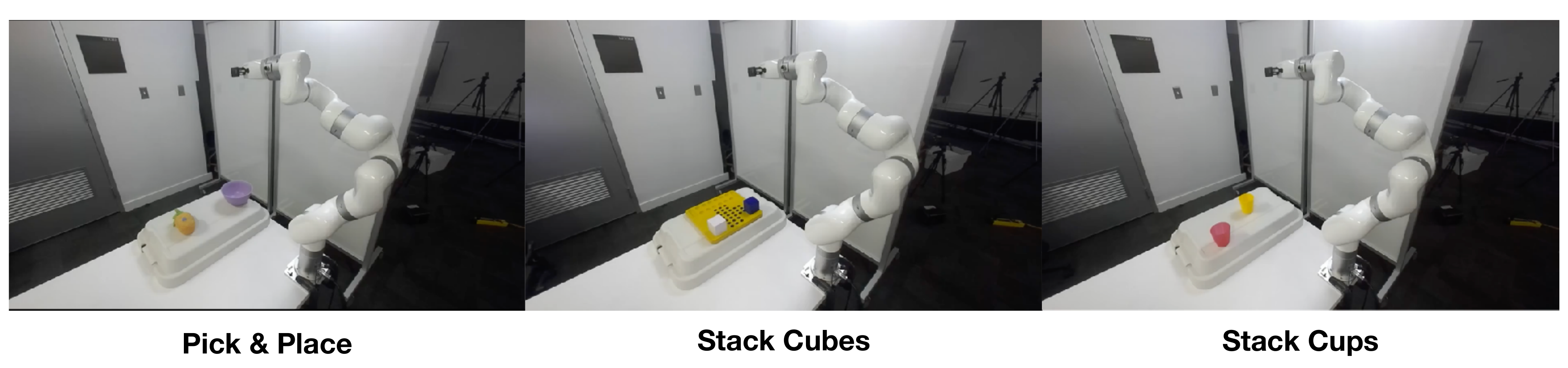}
  \caption{\small \textbf{xArm7 cross-embodiment setup.} The three evaluation tasks: pick-and-place, stack cubes, and stack cups. Test rollouts randomize initial object poses to assess generalization.}
  \label{fig:xarm_setup}
\end{figure}

The xArm7 evaluation uses a single ZED 2i stereo camera mounted to provide a frontal view of the workspace. We finetune all methods on 50 expert trajectories per task collected via human teleoperation and evaluate over 100 independent rollouts per task with \emph{randomized initial object poses}. Baselines are the most recent publicly available VLAs, $\pi_{0.5}$~\cite{pi05} and GR00T~N1.7~\cite{gr00t}. The three evaluation tasks are visualized in \cref{fig:xarm_setup}: pick-and-place, stack cubes, and stack cups.

\subsection{YAM Arm Setup}
\label{sec:appendix_yam}
\begin{figure}[!htbp]
    \centering
    \includegraphics[width=0.85\textwidth]{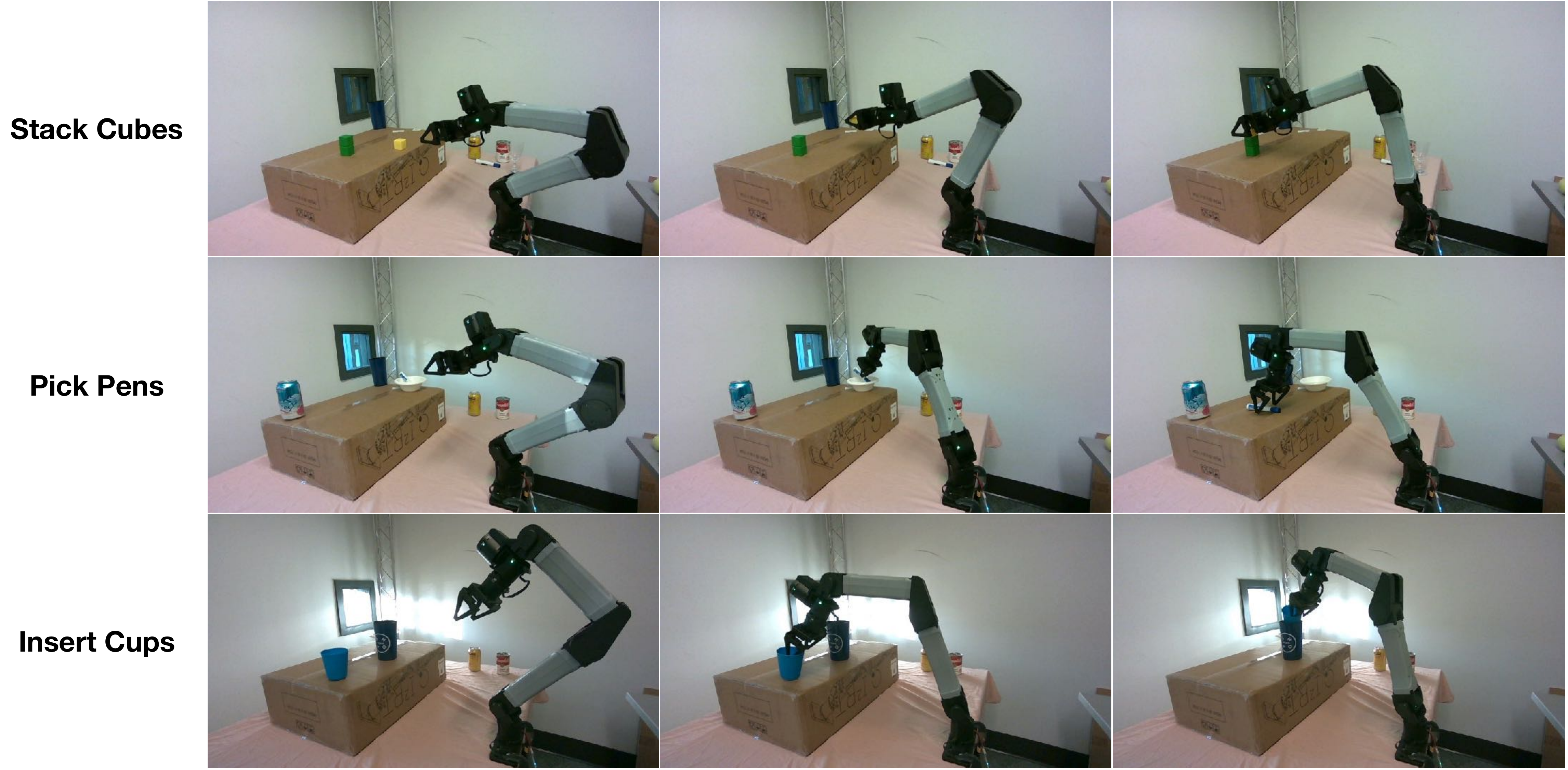}
    \caption{\textbf{YAM real-world tasks and setup.}}
    \label{fig:illu}
\end{figure}

The YAM evaluation uses an Intel RealSense D455 stereo camera and the same FoundationStereo~\cite{wen2025stereo} depth pipeline. Three manipulation tasks are evaluated:
\begin{itemize}
    \item \textbf{Stack Cubes:} the robot stacks a toy cube on top of two already-stacked cubes.
    \item \textbf{Pick Pens:} the robot picks a marker pen out of a container bowl.
    \item \textbf{Insert Cups:} the robot precisely inserts a paper cup into a round plastic tumbler.
\end{itemize}
We collect 20 expert trajectories per task via teleoperation. The joint RGB-XYZ model is post-trained for 500 additional steps on this dataset, and the point-action decoder is trained from scratch ($\sim$1~hour for 3{,}000 epochs on the combined three-task dataset). Baselines are $\pi_0$~\cite{pi0} and GR00T~N1.5~\cite{gr00t}. Each method is evaluated 20 times per task. The three tasks are illustrated in \cref{fig:illu}.

%% file: sections/appendix/D_qualitative_4d.tex
\section{Additional 4D Generation Analyses}
\label{sec:appendix_qual_4d}

This appendix complements \cref{tab:4dgen_quantitative} in the main paper with: (i) qualitative rollouts illustrating the temporal and geometric consistency of \methodname's joint RGB-XYZ prediction, (ii) head-to-head qualitative comparisons against 4D generation baselines, and (iii) a quantitative geometry-only evaluation against simulator ground-truth depth on out-of-distribution scenes.

\subsection{Temporal and Geometric Consistency}
\label{sec:appendix_qual_4d_consistency}
\begin{figure}[!htbp]
  \centering
  \includegraphics[width=0.85\textwidth]{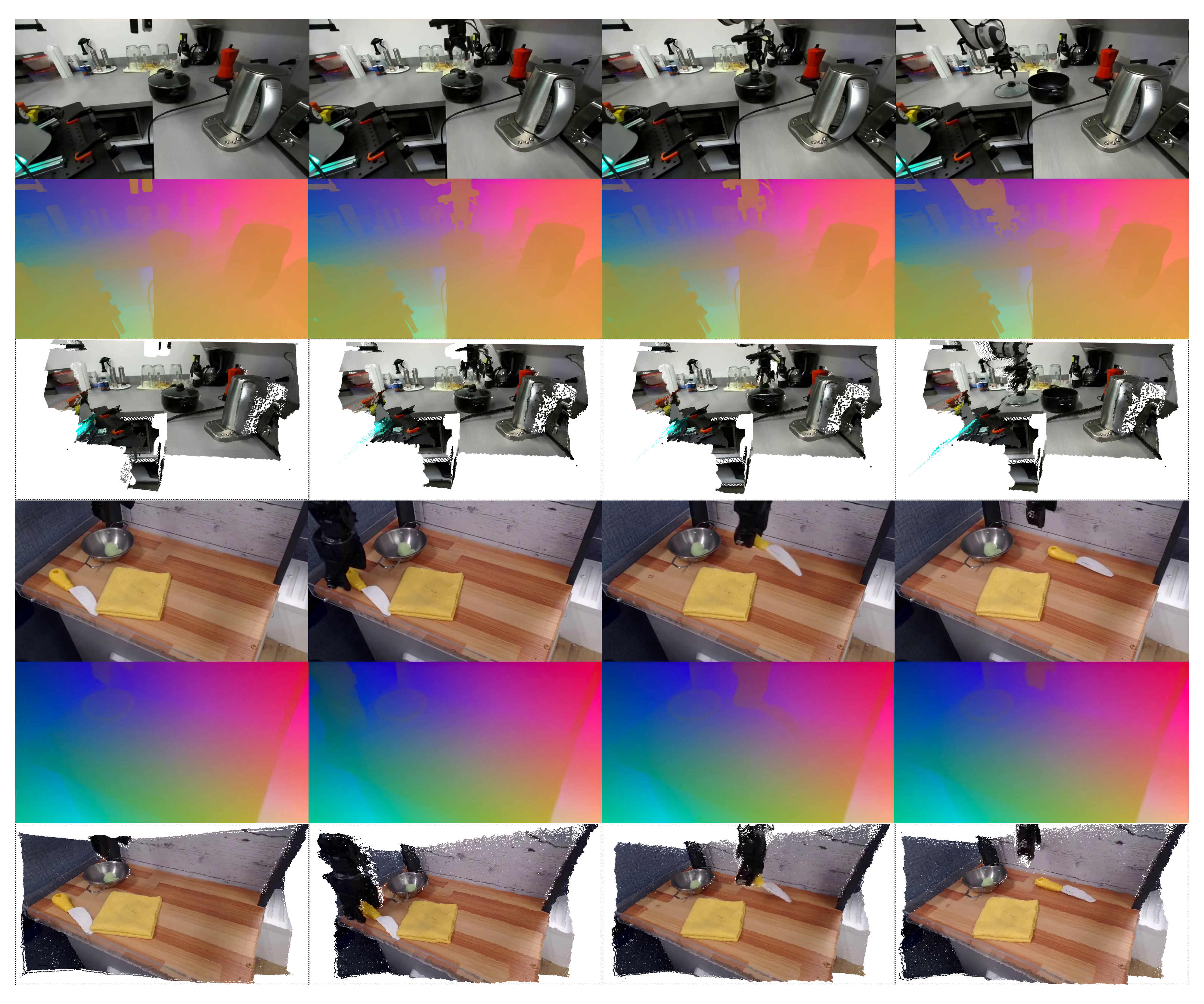}
  \caption{\small\textbf{Qualitative 4D generation by \methodname{} on real-world rollouts.} Top: a held-out trajectory from DROID~\cite{droid}; bottom: a held-out trajectory from BridgeData~V2~\cite{bridgev2}. Columns: four uniformly sampled timesteps. Rows: (a) generated RGB frames, (b) generated XYZ pointmaps, (c) rendered XYZ as a 3D point cloud.}
  \label{fig:4D_Qual}
\end{figure}

\cref{fig:4D_Qual} visualizes \methodname's predicted rollouts on two real-world scenes. Within each block, the four columns are four uniformly sampled timesteps of the predicted rollout (left to right), and the three rows show the jointly generated (a) RGB frames, (b) XYZ pointmaps, and (c) the XYZ rendered as a 3D point cloud. The figure highlights that the joint RGB-XYZ prediction remains coherent across time and across modalities---a prerequisite for the action decoder to extract reliable point dynamics.

\subsection{Comparison with 4D Generation Baselines}
\label{sec:appendix_qual_4d_compare}
\begin{figure}[!htbp]
  \centering
  \includegraphics[width=0.9\textwidth,height=0.85\textheight,keepaspectratio]{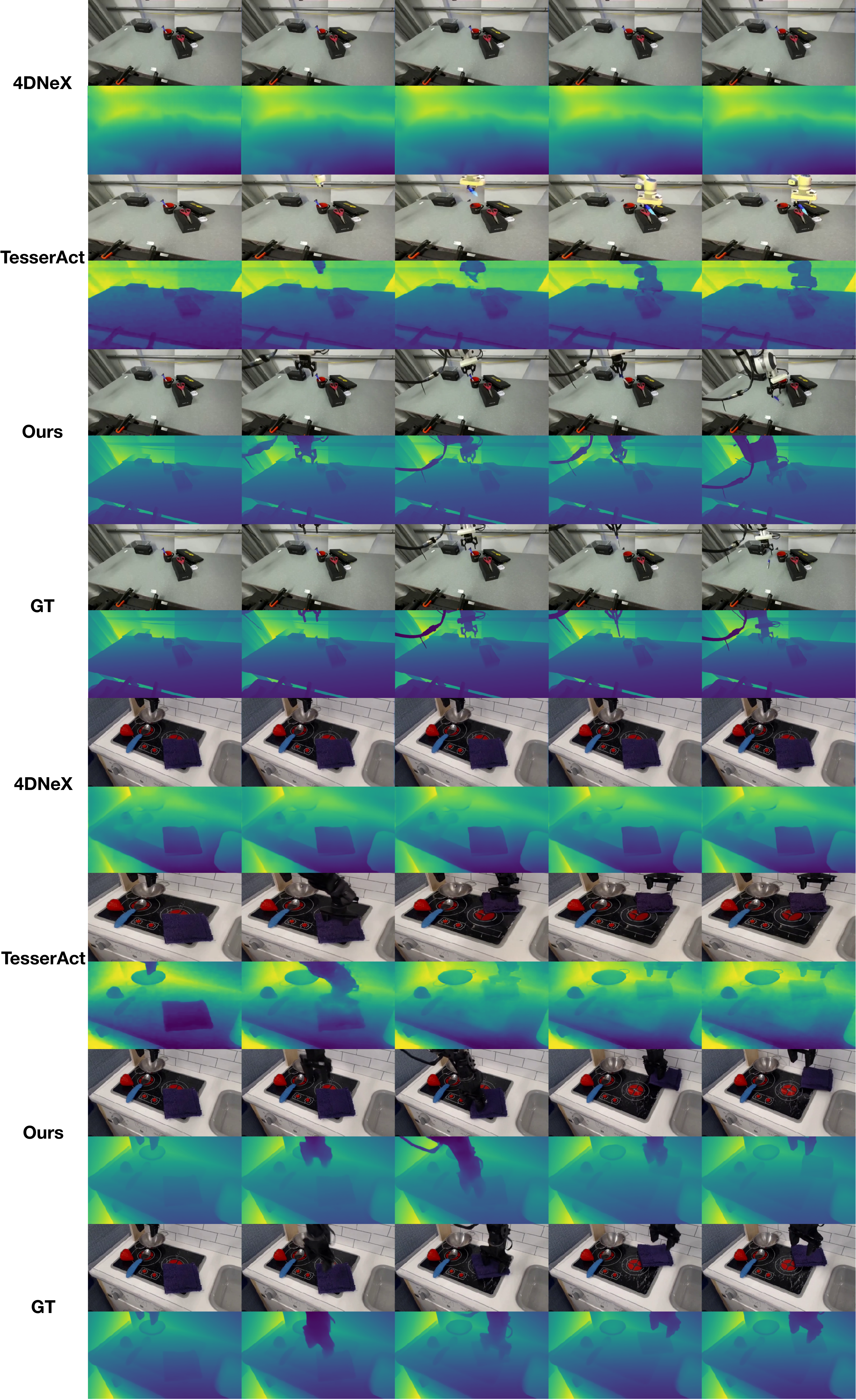}
  \caption{\small\textbf{Qualitative comparisons on RGB and depth generation.} Top: DROID~\cite{droid}. Bottom: BridgeData~V2~\cite{bridgev2}. \methodname (rightmost column group) produces sharper RGB and cleaner pixel-aligned depth than baselines.}
  \label{fig:appendix_qual4d}
\end{figure}

We additionally compare against state-of-the-art 4D generation baselines on the same two real-world domains (\cref{fig:appendix_qual4d}). \methodname delivers the highest visual quality in both RGB content fidelity and temporally consistent depth. In contrast, 4DNeX~\cite{4dnex} tends to predict overly static scenes with limited object motion, while TesserAct's~\cite{tesseract} depth maps exhibit noticeable noise and artifacts on real-world scenes. We attribute TesserAct's degradation to its training domain: although it includes BridgeData~V2 in pretraining, it is fine-tuned exclusively on simulation data for the Franka Panda robot, so it struggles to zero-shot generalize to the visually complex, real-world distributions in DROID.

\subsection{Out-of-Distribution Depth Reconstruction}
\label{sec:appendix_ood_depth}
To isolate the quality of the geometry channel from in-distribution training artifacts, we evaluate against simulator ground-truth depth on held-out RoboCasa365~\cite{robocasa365} trajectories that the 4D model has never seen during pretraining. \cref{tab:ood_recon} compares \methodname's directly generated XYZ pointmaps against state-of-the-art monocular (DepthAnything-V3) and dynamic-stereo (MegaSAM, the depth source used by LVP) depth estimators applied on top of our generated RGB. Our joint generation achieves the strongest performance on every metric with substantially lower variance, indicating that the geometry is genuinely learned by the joint model rather than recoverable post-hoc from the RGB output.
\input{table/ood_depth}

%% file: table/ood_depth.tex
\begin{table}[!htbp]
\centering
\caption{\textbf{OOD 4D reconstruction on RoboCasa365 against simulator ground-truth depth.} We compare \methodname's directly generated XYZ pointmaps against running state-of-the-art monocular and dynamic-stereo depth estimators (DepthAnything-V3, MegaSAM---the depth source used by LVP) on top of our generated RGB. Our joint generation achieves the best score on every metric with substantially lower variance.}
\label{tab:ood_recon}
\setlength{\tabcolsep}{10pt}
\renewcommand{\arraystretch}{1.05}
\begin{tabular}{l c c c c}
\toprule
Depth Source & AbsRel $\downarrow$ & $\delta_1$ $\uparrow$ & $\delta_2$ $\uparrow$ & Chamfer-$L_1$ $\downarrow$ \\
\midrule
MegaSAM (on our RGB)          & 0.187 & 0.775 & 0.877 & 0.327 \\
DepthAnything-V3 (on our RGB) & 0.198 & 0.755 & 0.867 & 0.361 \\
\midrule
\textbf{\methodname (joint RGB-XYZ)} & \textbf{0.118} & \textbf{0.872} & \textbf{0.904} & \textbf{0.151} \\
\bottomrule
\end{tabular}
\end{table}

%% file: sections/appendix/E_failure_modes.tex
\section{Failure Mode Analysis}
\label{sec:appendix_failure}

We attribute the failures observed during deployment to two dominant causes:

\noindent\textbf{(1) Self-occlusion.} When the robot occludes parts of itself in the generated video, the robot-centric XYZ pointmap is incomplete, and the action decoder receives degraded geometric input. This is most pronounced for end-effector poses that are partially behind the gripper from the camera's vantage point.

\noindent\textbf{(2) Open-loop execution.} Because the policy executes a 49-step action chunk in a single forward pass without re-planning, it cannot recover from unexpected events such as objects slipping during grasping or perturbations from environmental contact. Closing this loop---e.g., by distilling the video backbone into an autoregressive, causally cached architecture~\cite{huang2025self,causvid,zhu2026causal}---is a clear direction for future work.